\documentclass[twoside]{article}
\usepackage{PRIMEarxiv}
\usepackage[utf8]{inputenc}
\usepackage[T1]{fontenc}
\usepackage[colorlinks=true, urlcolor=blue, citecolor=black]{hyperref}

\usepackage{todonotes}
\usepackage{graphicx}
\usepackage{caption}
\usepackage{subcaption}
\usepackage{float}
\usepackage{listings}
\usepackage{multirow}
\usepackage{amssymb}
\usepackage{mathtools}
\usepackage{amsmath}
\usepackage{comment}
\usepackage{algorithm}
\usepackage{algorithmicx}%
\usepackage{algpseudocode}
\usepackage{siunitx}
\usepackage{dsfont}
\usepackage{xcolor, colortbl}
\usepackage{tikz}
\usetikzlibrary{fit}
\usetikzlibrary{positioning}
\usetikzlibrary{shapes.geometric}
\usepackage{rotating}
\usepackage{cite}
\usepackage{enumitem}
\usepackage{booktabs}

\makeatletter
\def\hlineb#1{%
	\noalign{\ifnum0=`}\fi\hrule \@height #1 %
	\futurelet\reserved@a\@xhline}
\makeatother

\title{Deep Clustering Using the Soft Silhouette Score: Towards Compact and Well-Separated Clusters}

\author{
  Georgios Vardakas \\
  \textit{Department of Computer Science and Engineering} \\
  University of Ioannina \\
  GR 45110, Ioannina, Greece \\
  \texttt{g.vardakas@uoi.gr}
  \And
  Ioannis Papakostas  \\
  \textit{Department of Computer Science and Engineering} \\
  University of Ioannina \\
  GR 45110, Ioannina, Greece \\
  \texttt{john.s.papakostas@gmail.com}
  \And
  Aristidis Likas \\
  \textit{Department of Computer Science and Engineering} \\
  University of Ioannina \\
  GR 45110, Ioannina, Greece \\
  \texttt{arly@cs.uoi.gr}
}

\begin{document}
\maketitle              

\begin{abstract}
Unsupervised learning has gained prominence in the big data era, offering a means to extract valuable insights from unlabeled datasets. Deep clustering has emerged as an important unsupervised category, aiming to exploit the non-linear mapping capabilities of neural networks in order to enhance clustering performance. The majority of deep clustering literature focuses on minimizing the inner-cluster variability in some embedded space while keeping the learned representation consistent with the original high-dimensional dataset. In this work, we propose \emph{soft silhoutte}, a probabilistic formulation of the silhouette coefficient. Soft silhouette rewards compact and distinctly separated clustering solutions like the conventional silhouette coefficient. When optimized within a deep clustering framework, soft silhouette guides the learned representations towards forming compact and well-separated clusters. In addition, we introduce an autoencoder-based deep learning architecture that is suitable for optimizing the soft silhouette objective function. The proposed deep clustering method has been tested and compared with several well-studied deep clustering methods on various benchmark datasets, yielding very satisfactory clustering results.
 
\end{abstract}
	
\section{Introduction}
Unsupervised learning has become increasingly important due to the rise of big data collection and the high cost associated with acquiring labeled data. This field of research encompasses various techniques, some of which include generative models~\cite{harshvardhan2020comprehensive}, representation learning, dimensionality reduction~\cite{zhang2018network} and clustering~\cite{nasraoui2019clustering}. Such methods enable us to extract meaningful insight on properties of the data, without relying on explicit guidance or supervision from pre-existing labels. 
Clustering is a fundamental unsupervised learning task with numerous applications in computer science and many other scientific fields~\cite{filippone2008survey, jain2010data, ezugwu2022comprehensive}. Even though a strict definition of clustering may be challenging to establish, a more flexible interpretation can be stated as follows: Clustering is the process of partitioning a set of objects into groups, known as clusters, such that data in the same group share ``common'' characteristics while ``differing'' from data in other groups. While the above clustering definition is simple, it is proven to be a hard machine learning problem~\cite{aloise2009np}. More specifically, it is known that its difficulty arises from several factors like data prepossessing and representation, clustering criterion, optimization algorithm and parameter initialization.

Due to its particular importance, clustering is a well-studied problem with numerous proposed approaches. Generally, they can be classified as hierarchical (divisive or agglomerative), model-based (e.g. $k$-means~\cite{macqueen1967some}, mixture models~\cite{bishop2006pattern}) and density-based (e.g. DBSCAN~\cite{ester1996density}, DensityPeaks~\cite{rodriguez2014clustering}). Most methods are effective when the data space is low dimensional and not complex. Various feature extraction and feature transformation methods have been proposed to map the original complex data to a simpler feature space as a prepossessing step to address those limitations. Some of the methods include Principal Component Analysis~\cite{wold1987principal}, Non-negative Matrix Factorization~\cite{lee1999learning}, Spectral methods~\cite{ng2002spectral}, and Minimum Density Hyperplanes~\cite{pavlidis2016minimum}. 

More recently, deep neural networks (DNNs) have been employed for clustering in the context of deep learning. DNNs are used to learn rich and useful data representations from data collections without heavily relying on human-engineered features~\cite{bengio2013representation}. They can improve the performance of both supervised and unsupervised learning tasks because of their excellent nonlinear mapping capability and flexibility~\cite{krizhevsky2012imagenet,lecun2015deep}. Although clustering has not initially been the primary goal of deep learning, several clustering methods have been proposed that exploit the representational power of neural networks; thus, the deep clustering category of methods has emerged. Such methods aim to improve the quality of clustering results by appropriately training neural networks to transform the input data and generate \emph{cluster-friendly} representations, meaning that in the latent space the data will form compact and, in the optimal case, well-separated clusters~\cite{aljalbout2018clustering, min2018survey, inbook, zhou2022comprehensive, ren2022deep}. 

Several DNN models have been utilized in the deep clustering framework~\cite{ren2022deep}. More specifically, popular architectures include Generative Adversarial Networks~\cite{goodfellow2014generative}, Variational Autoencoder~\cite{kingma2013auto}, Graph Neural Networks~\cite{zhou2020graph} and regular DNNs have been utilized by methods such as ClusterGan~\cite{mukherjee2019clustergan}, VaDE~\cite{jiang2016variational}, JULE~\cite{yang2016joint}, NIMLC~\cite{vardakas2023neural}. 

However, the majority of the methods rely on training Autoencoders (AE), which is the most utilized deep learning model for clustering. AE-based deep clustering methodologies attempt to exploit the non-linear capabilities of the encoder and decoder models in order to assist in latent space~\cite{yang2017towards}. To achieve this goal, novel objective functions have been proposed that integrate the typical AE reconstruction error with a clustering loss in order to train the AE network so that in the learned embedded space, the data will form more compact clusters (achieved through minimization of a clustering objective), while at the same time retaining the information of the original data (achieved by minimizing the AE reconstruction error).  

As presented in the related work section, the vast majority of AE-based methods learn a representation in which individual clusters have small inner cluster variability. Most common approaches are the minimization of the $k$-means error, or the $KL$ divergence between the soft clustering assignments and a target distribution. Such a representation has been shown to improve the clustering results in several scenarios. However, minimizing only the inner cluster distance is a suboptimal strategy. Our motivation is to formulate a deep clustering objective that simultaneously considers both the inner cluster distance and the outer cluster separation. This is achieved by optimizing the \emph{soft silhouette} objective introduced in this work.

Assessing the quality of a clustering solution is typically a challenging task. In this direction, several quality measures have been proposed which can be categorized as external and internal measures~\cite{rendon2011internal}. External quality measures, as the name suggests, use additional information about the data as the ground truth labels. Well-known  external evaluation measures include Normalized Mutual Information (NMI)~\cite{estevez2009normalized}, Adjusted Mutual Information (AMI)~\cite{JMLR:v11:vinh10a}, Adjusted Rand Index and (ARI)~\cite{hubert1985comparing,chacon2022minimum}. However, such measures are not applicable in real-world applications where the ground truth labels are absent. Internal quality measures, on the other hand, can be applied to the clustering problem since they are based solely on the information intrinsic to the data. Some typical internal clustering measures that take into account both cluster compactness and separation are the Dunn index~\cite{dunn1973fuzzy}, the Calinski-Harabasz index~\cite{calinski1974dendrite}, the Davies-Bouldin index~\cite{davies1979cluster}, and the silhouette~\cite{rousseeuw1987silhouettes}. In particular, the silhouette coefficient is the most widely used and successful internal validation measure~\cite{arbelaitz2013extensive}.

The typical silhouette is considered as an effective clustering quality measure that combines both inter and intra cluster information. Specifically, silhouette rewards clustering solutions that exhibit both compactness within individual clusters and clear separation between clusters. However, it assumes a hard clustering solution, thus it cannot be used to evaluate probabilistic clustering solutions, unless they are transformed to discrete ones based on maximum cluster membership probability. In addition, the silhouette score cannot be efficiently used as a clustering objective for neural network training since it is not differentiable. 

In this work, in order to overcome the above limitations, we propose an extension of the silhouette score, called \emph{soft silhouette} score, that evaluates the quality of probabilistic clustering solutions without requiring their transformation to discrete ones. Besides this obvious advantage, a notable property of soft silhouette is that it is differentiable with respect to cluster assignment probabilities. Assuming that such probabilities are provided by a parametric machine learning model, the soft silhouette score be used as a clustering objective function to train parametric probabilistic models using typical gradient-based approaches.

To this end, we propose a novel AE-based  deep clustering methodology that directly provides cluster assignment probabilities as network outputs and exploits the soft silhouette score as a clustering objective. In this way, by training the network using soft silhouette, we achieve minimization of the inter-cluster variance, while at the same time maximizing the margin between clusters in the embedded space.


The rest of the paper is organized as follows. In Section~\ref{sec:RelatedWork}, we give a brief overview of the deep clustering methods based on the AE model, while in Section~\ref{sec:SS} the soft silhouette score introduced. Then in Section~\ref{sec:Method} we describe the proposed deep clustering methodology by presenting the model architecture, the corresponding objective function as well as the training method. Finally, in Section~\ref{sec:Experiments}, we provide extensive experimental results and comparisons, while in Section~\ref{sec:Conclusion}, we provide conclusions and future work suggestions.

\section{Deep Clustering using Autoencoders}

\label{sec:RelatedWork}

The autoencoder (AE) constitutes the most widely used model in deep clustering methodologies~\cite{ren2022deep}. It consists of an encoder network $z=f_\theta (x)$ which given an input $x$ provides emdedding $z$, and the decoder network $\hat{x}=g_w (z)$ that provides the reconstruction $\hat{x}$ given the embedding $z$. In the typical autoencoder case, given a dataset $X=\{x_1, \ldots, x_N\}$, the parameters $\theta$ and $w$ are adjusted by minimizing the reconstruction loss:
\begin{equation}
\label{eq:Rec-loss}
\mathcal{L}_{rec}=\frac{1}{N}\sum\limits_{i=1}^{N} ||x_{i} - g_\theta(f_w(x_i))||^2
\end{equation}
A straightforward approach for AE-based clustering is to first train the AE using the reconstruction loss and then cluster the embeddings $z_i = f_w(x_i)$ using any clustering method. Thus data projection and clustering are performed independently. However, it has been found the better results are obtained if the embeddings are formed taking into account both reconstruction and clustering.

Therefore, the \emph{AE clustering framework} has emerged, where the goal is to create cluster-friendly embeddings $z_i$. To achieve this goal, the reconstruction loss is enhanced with an appropriately defined clustering loss $\mathcal{L}_{cl}$ resulting  in a total loss of the form:
\begin{equation}
\label{eq:AE-loss}
\mathcal{L}_{AE} = \mathcal{L}_{rec} + \lambda\mathcal{L}_{cl},
\end{equation}
where the hyperparameter $\lambda$ balances the relative importance the two objectives. 

It should be noted that the of minimization clustering loss enforces the formation of embeddings $z_i$ with small cluster variance. An easily obtained trivial solution exists, where all data points $x_i$ are mapped to embeddings $z_i$ that are all very close to each other (ie. the encoder is actual a constant function). To avoid this trivial solution the reconstruction loss is added that forces the embeddings $z_i$ to retain the information of the original dataset. In essence, AE clustering methods strive to create embeddings that form compact clusters, while keeping the characteristics of the original dataset. The most widely used AE-based methods are summarized next.

Inspired by the $t$-SNE~\cite{van2008visualizing} algorithm, the Deep Embedding Clustering (DEC)~\cite{xie2016unsupervised} method has been proposed that optimizes both the reconstruction objective and a clustering objective. DEC transforms the data in the embedded space using an autoencoder and then optimizes a clustering loss defined by the $KL$ divergence between two distributions $p_{ij}$ and $q_{ij}$: $q_{ij}$ are soft clustering assignments of the data based on the distances in the embedded space between data points and cluster centers, and $p_{ij}$ is an adjusted target distribution aiming to enhance the clustering quality by leveraging the soft cluster assignments. More specifically, $q_{ij}$ is defined as 
\begin{equation}
\label{eq:qij}
q_{ij} = \frac{(1 + ||z_i - \mu_j||^2/\alpha)^{-\frac{a+1}{2}}}{\sum_{j'}(1 + ||z_i - \mu_{j'}||^2/\alpha)^{-\frac{a+1}{2}}},
\end{equation}
where $z_i = f_w(x_i)$, $\mu_j$ is a cluster center in the embedded space and $\alpha$ is the degrees of freedom. Additionally, $p_{ij} = \frac{q_{ij}^2/f_{ij}}{\sum_{j'}q_{ij}^2/f_{ij}}$ (with $f_{ij} = \sum_i q_{ij}$) is the target probability distribution that aims to sharpen the cluster probability assignments.

 A modification of this method is the Improved Deep Clustering with local structure preservation (IDEC)~\cite{guo2017improved}. Both DEC and IDEC optimize the same objective function:
\begin{equation}
\label{eq:IDEC}
\mathcal{L}_{rec} + \lambda\sum\limits_{i=1}^{n}\sum\limits_{j=1}^{k} p_{ij}\log{\frac{p_{ij}}{q_{ij}}}.
\end{equation}
Specifically, DEC performs pretraining to minimize the reconstruction loss and subsequently excludes the decoder part of the network, focusing solely on the clustering loss during the training phase. In contrast, IDEC simultaneously optimizes both the reconstruction loss and the clustering loss during the training phase of the AE. From eq.~\ref{eq:qij} and eq.~\ref{eq:IDEC} it is clear that the above approaches aim to minimize only an inner cluster distance loss.

Similar to DEC, the Deep Clustering Network (DCN)~\cite{yang2017towards} jointly learns the embeddings and the cluster assignments by directly optimizing the $k$-means clustering loss on the embedded space. The optimized objective function is:
\begin{equation}
\label{eq:DCNLoss}
\mathcal{L}_{rec} + \lambda \sum_{i=1}^{n} ||z_i - M s_i||^2
\end{equation}
where $z_i = f_w(x_i)$, $M$ is a matrix containing the $k$ cluster centers in the embedded space, and $s_i$ is the cluster assignment vector for data point $x_i$ with only one non-zero element. An analogous early work is Autoencoder-based data clustering (AEC)~\cite{song2013auto} which also aims to minimize the distance between embedded data and their nearest cluster centers.




\section{The soft silhouette score}
\label{sec:SS}
\subsection{Silhouette}
The silhouette score~\cite{rousseeuw1987silhouettes} is a measure utilized to assess the quality of a clustering solution. It assumes that a good clustering solution encompasses compact and well-separated clusters. Assume we are given a partition $C=\{C_1,...,C_K\}$ of a dataset $X=\{x_1, \ldots, x_N\}$ into $K$ clusters. Let also $d(x_i,x_j)$ denote the distance between $x_i$ and $x_j$. 

The silhouette score computation proceeds by evaluating the individual silhouette score $s(x_i)$ of each data point $x_i$ as follows. We first compute its average distance $a(x_i)$ to all other data points within its cluster $C_I$:
\begin{equation}
\label{eq:ai}
a(x_i) = \frac{1}{|C_I| - 1} \sum_{x_j \in C_I, i \neq j} d(x_i, x_j),
\end{equation}
where $|C_I|$ represents the cardinality of cluster $C_I$, where $|C_I| > 1$. The value of $a(x_i)$ value quantifies how well the datapoint $x_i$ fits within its cluster. A low value of $a(x_i)$ indicates that $x_i$ is similar to its cluster members, suggesting that $x_i$ is probably grouped correctly. Conversely, a higher value of $a(x_i)$ indicates that $x_i$ is far from its cluster members.

The silhouette score also requires the calculation of the minimum average outer-cluster distance $b(x_i)$ for each datapoint $x_i \in C_I$ defined as
\begin{equation}
\label{eq:bi}
b(x_i) = \min_{J \neq I}\frac{1}{|C_J|} \sum_{x_j \in C_J} d(x_i, x_j).
\end{equation}
A large $b(x_i)$ value indicates that the datapoint $x_i$ significantly differs from datapoints in other clusters which is desirable. 

The silhouette score of a datapoint $x_i$ takes into account the requirements for small $a(x_i)$ and large $b(x_i)$ and is defined as:
\begin{equation}
\label{eq:si}
s(x_i) = \frac{b(x_i) - a(x_i)}{\max\left\{a(x_i), b(x_i)\right\}}.
\end{equation}
It should be noted that $ -1\leq s(x_i)\leq 1$. A value close to $1$ is achieved when $a(x_i)$ is small and $b(x_i)$ is high. This indicates that $x_i$ has been assigned to a compact, well-separated cluster. In contrast, a value close to $-1$ suggests that the $x_i$ is more similar to points in other clusters than to points in its cluster, thus it has been probably assigned a wrong cluster label. 

The total silhouette score for the whole partition $C$ of the dataset $X$ is obtained by aggregating the individual silhouette values through typical averaging:
\begin{equation}
\label{eq:Silhouette}
S(X) = \frac{1}{N} \sum_{i=1}^{N} s(x_i)    
\end{equation}

The silhouette score~\cite{rousseeuw1987silhouettes} is not only suitable for (internal) clustering evaluation but also defines an intuitive clustering objective that rewards compact and well-separated clusters. As presented in related work, while several deep clustering objectives aim to provide compact clustering solutions, they  do not optimize explicitly for cluster separability. Next, we introduce a probabilistic silhouette score, termed soft silhouette, which allows us to optimize for both compact and well-separated clusters.

\subsection{Soft Silhouette}
The soft silhouette score introduced below constitutes an extension of the typical silhouette score that assumes probabilistic cluster assignments instead of hard cluster assignments. More specifically, assume a dataset $X=\{x_1, \ldots, x_N\}$ partitioned into $K$ clusters $C=\{C_1, \ldots, C_K\}$ and let $d(x_i,x_j)$ the distance between data points $x_i$ and $x_j$. Let also $P_{C_I}(x_i)$ denote the probability that $x_i$ belongs to cluster $C_I$. Obviously $\sum_{I=1}^K P_{C_I}(x_i)=1$.

Assuming that $x_i$ belongs to cluster $C_I$ we define as:
\begin{itemize}
\item $a_{C_I}(x_i)$ the value of the distance of $x_i$ to cluster $C_I$. This is actually a weighted average (expected value) of the distances of $x_i$ to all other points $x_j \in X$ with weight the probability $P_{C_I}(x_j)$ (ie. that $x_j$ belongs to the cluster of interest $C_I$).
\begin{equation}
\label{eq:soft_ai}
a_{C_I}(x_i) = \frac{\sum\limits_{j=1}^{N} P_{C_I}(x_j) d(x_i, x_j)}{\sum\limits_{j=1, j \neq i}^{N} P_{C_I}(x_j)}  
\end{equation}

\item $b_{C_I}(x_i)$ the minimum value of the (expected) distance of $x_i$ from the other clusters $C_j$ different from $C_I$.  
\begin{equation}
\label{eq:soft_bi}
b_{C_I}(x_i) = \min_{J \neq I} \frac{\sum\limits_{j=1}^{N} P_{C_J}(x_j) d(x_i, x_j)}{\sum\limits_{j=1, j \neq i}^{N} P_{C_J}(x_j)}  = \min_{J \neq I} a_{C_J}(x_i)
\end{equation}
\item $s_{C_I}(x_i)$ the conditional silhouette value for $x_i$ given that it belongs to cluster $C_I$: 
\begin{equation}
\label{eq:soft_si}
s_{C_I}(x_i) = \frac{b_{C_I}(x_i) - a_{C_I}(x_i)}{\max\left\{a_{C_I}(x_i), b_{C_I}(x_i)\right\}}  
\end{equation}
\end{itemize}
Then the soft silhouette value $sf(x_i)$ of data point $x_i$ is computed as the expected value of $s_{C_I}(x_i)$ with respect to its cluster assignment probabilities $P_{C_I}(x_i)$:  
\begin{equation}
\label{eq:soft_Sil_xi}
sf(x_i) = \sum\limits_{i = I}^{K} P_{C_I}(x_i) s_{C_I}(x_i)
\end{equation}
and the total soft silhouette score $Sf(X)$ of the partition is computed by aggregating (averaging) the individual scores $sf(x_i)$:
\begin{equation}
\label{eq:SoftSilhouette}
Sf(X) = \frac{1}{N} \sum_{i=1}^{N} sf(x_i),
\end{equation}
It should be noted that in the case of hard clustering, the cluster assignment probability vectors become one-hot vectors and the soft silhouette equations become similar to the typical silhouette equations. 

It is obvious from the above equations that soft silhouette is differentiable with respect to the cluster assignment probabilities. Therefore it can can be employed as a clustering objective function to be optimized in a deep learning framework. The major advantage of this objective is that it optimizes simultaneously both cluster compactness and separation. Such a deep clustering approach is presented next.

\section{The DCSS method: Deep Clustering using Soft Silhouette}
\label{sec:Method}
In this section we propose the Deep Clustering using Soft Silhouette (DCSS) algorithm which belongs the category of AE-based deep clustering methods that employs soft silhouette as a clustering loss. 

As described in Section~\ref{sec:RelatedWork} a typical AE-based deep clustering method employs an encoder network \begin{equation*} z = f_w(x), \qquad
f_w(\cdot): \mathbb{R}^d \rightarrow \mathbb{R}^m,
\end{equation*} that provides the latent representations (embeddings) $z$ and a decoder network \begin{equation*} \hat{x} = g_\theta(x), \qquad
g_\theta(\cdot): \mathbb{R}^m \rightarrow \mathbb{R}^d,
\end{equation*} that reconstructs the outputs given the embeddings. The networks are trained to optimize a total loss that is the sum of the reconstruction loss and the clustering loss: $\mathcal{L}_{AE} = \mathcal{L}_{rec} + \lambda\mathcal{L}_{cl}$. 

\begin{figure}[ht]
\centering
\begin{tikzpicture}[
	embedded/.style={rectangle, draw=black, fill=violet!40, thick, minimum width=15pt, minimum height=15mm},
 	cluster/.style={rectangle, draw=black, fill=red!40, thick, minimum width=15pt, minimum height=15mm},
    sclustering/.style={rectangle, draw=black, fill=yellow!40, thick, minimum width=15pt, minimum height=15mm},
	generator/.style={trapezium, trapezium angle=60, fill=blue!40, rotate=0, shape border rotate=90, minimum width=35mm, minimum height=10mm, trapezium stretches = true, draw, thick},
	image/.style={rectangle, draw=black, fill=gray!25, thick, minimum width=15pt, minimum height=35mm},
	encoder/.style={trapezium, trapezium angle=120, fill=green!40, rotate=0, shape border rotate=90, minimum width=35mm, minimum height=10mm, trapezium stretches = true, draw, thick},
	]
	\def\horizontalspace{1.0cm}
    \def\verticalspace{1.75cm}

 	\node[image]		(image)	                                            {$X$};
    \node[encoder]		(encoder)	[right = \horizontalspace of image]		{$f_{w}$};
    \node[embedded]     (embedded)  [right = \horizontalspace of encoder]   {$Z$};
	\node[generator]	(generator) [right =  \horizontalspace of embedded]	{$g_{\theta}$};
	\node[image]		(himage)	[right = \horizontalspace of generator] {$\hat{X}$};
	\node[cluster]      (cluster)   [below = \verticalspace of embedded]    {RBF};
    \node[cluster]      (softmax)   [right = \horizontalspace of cluster]    {\rotatebox{90}{Softmax}};
    \node[sclustering]   (sclustering)   [right = 1.3cm of softmax]  {\rotatebox{90}{Clustering}};
    
    \draw[->, line width=1] (image) -- (encoder);
    \draw[->, line width=1] (encoder) -- (embedded);
    \draw[->, line width=1] (embedded) -- (generator);
    \draw[->, line width=1] (generator) -- (himage);
    \draw[->, line width=1] (embedded) -- (cluster);
    \draw[->, line width=1] (cluster) -- (softmax);
    \draw[->, line width=1] (softmax) -- (sclustering);

	\node[draw=black, dotted, line width=1pt, inner sep=17.5pt, fit=(encoder) (embedded) (generator)] (ae_rect) {};
	\node[above] at (ae_rect.north) {Autoencoder};

    \node[draw=black, dotted, line width=1pt, inner sep=10pt, fit=(cluster) (softmax)] (cluster_rect) {};
	\node[below] at (cluster_rect.south) {Clustering Network ($h_r$)};
    
	\end{tikzpicture}
	\caption{The proposed model architecture.}
	\label{fig:ClAutoencoder}
\end{figure}
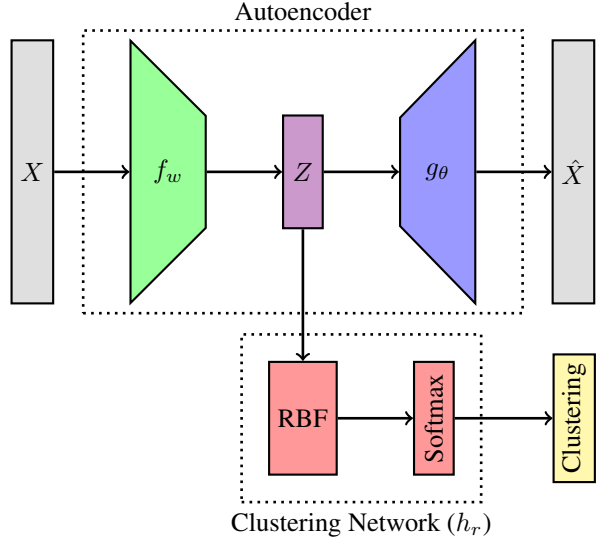

In the proposed approach the clustering loss will be based on the soft silhouette score, which requires the cluster assignment probabilities $p(x)=(p_1(x),\ldots, p_K(x))$ for an input $x$. For this reason, we enrich the AE-model with an additional network $h_r(z)$, called clustering network, that takes as input the embedding $z=f_w (x)$ of a data point $x$ and outputs the cluster assignment probabilities $p_j (x)={h_r}_j (x)$ for $j=1,\ldots, K$. Therefore, given the dataset $X=\{x_1, \ldots, x_N\}$, the embedding $z_i=f_w(x_i)$ are first computed. Then the pairwise distances $d(z_i, z_j)$ and the cluster assignment probability vectors $p(x_i)=h_r(z_i)$ are specified, required for the soft silhouette computation. 

The proposed three network architecture is illustrated in Fig.~\ref{fig:ClAutoencoder}. It can be observed that the clustering network operates in parallel with the decoder network. For an input vector $x$ the model provides the embedding vector $z$, the reconstruction vector $\hat{x}$ and the probability vector $p(x)$. 

Based on the experimentation with several alternatives, we have selected as clustering network $h_r$ an  a Radial Basis Function (RBF)~\cite{buhmann2000radial} model with a softmax output unit that provides the probability vector of the cluster assignments. The number of RBF units is set equal to the number of clusters $K$.



Soft silhouette is a criterion that should maximized in order to obtain solutions of good quality. Since a clustering loss is a quantity to be minimized, we take into account that $Sf\leq 1$ and define the clustering loss as follows:
\begin{equation}
\label{eq:cluster_loss}
\mathcal{L}_{cl} = 1 - Sf
\end{equation}
Note that $L_{cl}$ is always positive and attains each minimum value when $Sf$ is maximum ($Sf=1$). Thus the total loss for model training is specialized as follows:
\begin{equation}
\label{eq:rec_loss}
L_{AE} = \frac{1}{N} \sum_{i=1}^{N} ||x_i - g_{\theta}(f_{w}(x_i))||^{2} + \lambda (1-Sf(h_r (X)))
\end{equation}
where $h_r(X)=\{h_r (x_1), \ldots, h_r(x_N)\}$ are the cluster assignment probability vectors. It should be noted that  the pairwise distances $d(f_w(x_i), f_w (x_j))$ between the embeddings are also involved in the $Sf$ computation. In this work the Euclidean distance has been used. Since $L_{AE}$ is differentiable with respect to the model parameters $w, \theta, r$ it can be minimized using typical gradient-based procedures. 

A technical issue that has emerged when training the model is that in many cases a trivial solution is attained where the output probabilities tend to be uniform (ie. equal to $1/K$) for many data points. To overcome this difficulty, we have included an additional term to the objective function that penalizes uniform solutions by minimizing the entropy of the output probability vectors. 
Thus, the entropy regularization term $L_{reg}$ is defined as follows: 
\begin{equation}
\label{eq:Entropy}
H(h_r (X)) = -\sum\limits_{i=1}^{N} \sum\limits_{j=1}^{K} {h_r}_j (x_i)\log{{h_r}_j (x_i)}
\end{equation}

The final total loss that is minimized to train our model is:
\begin{equation}
\label{eq:Autoencoder_optimization}
\mathcal{L}_{AE} = \mathcal{L}_{rec} + \lambda_{1} \mathcal{L}_{cl} + \lambda_{2} \mathcal{L}_{reg}
\end{equation}
and in more detail
\begin{equation}
\mathcal{L}_{AE} = \frac{1}{N} \sum_{i=1}^{N} ||x_i - g_{\theta}(f_{w}(x_i))||^{2} + \lambda_1 (1-Sf(h_r (X))) - \lambda_2 \frac{1}{N}\sum\limits_{i=1}^{N} \sum\limits_{j=1}^{K} {h_r}_j (x_i)\log{{h_r}_j (x_i)}
\end{equation}


The details of approach, called Deep Clustering using Soft Silhouette (DCSS), are summarized in Algorithm~\ref{alg:method}.

\begin{algorithm}[t]
\begin{algorithmic}[1]
\Require $X$ (dataset)
\Require $K$ (number of clusters)
\Require $\lambda_1$, $\lambda_2$ (regularization hyperparameters)
\vspace{0.2em}
\hrule
\vspace{0.2em}
\State Randomly initialize the $w$ and $\theta$ AE parameters.
\item[] \textbf{Stage 1: Pretraining}
\State Pretrain the encoder $f_w$ and decoder $g_{\theta}$ by minimizing the reconstruction error $\mathcal{L}_{rec}$ (eq.~\ref{eq:Rec-loss}) through gradient based optimization for $T_{pr}$ epochs. \Comment{We employed batch training using the Adam optimizer.}
\State Apply $k$-means with $K$ clusters to the learned representations $z = f_w(X)$.
\State Initialize the parameters of the clustering network $h_r$ using the $k$-means result.
\item[] \textbf{Stage 2: Training}
\State Update the parameters $\theta$, $w$ and $r$ by minimizing the total loss (eq.~\ref{eq:Autoencoder_optimization}) until convergence through gradient based optimization to obtain $\theta^{\star}$, $w^{\star}$ and $r^{\star}$. \Comment{We employed batch training using the Adam optimizer.}
\item[] \textbf{Stage 3: Inference}
\State Compute the clustering solution $C = \arg\max softmax( h_{r^{\star}}(f_{w^{\star}}(X)))$. \Comment{Data clustering.}
\State \Return the clustering solution $C$ and the learned parameters $w^{\star}$, $\theta^{\star}$, $r^{\star}$.
\end{algorithmic}
\caption{Deep Clustering using Soft Silhouette algorithm (DCSS)}
\label{alg:method}
\end{algorithm}

\section{Experiments}
\label{sec:Experiments}
In this section we present our experimental results on both real datasets and a synthetic dataset. In the first part, we demonstrate the representation learning capabilities of several methods compared to DCSS on a synthetic dataset. In the second part, we demonstrate the deep clustering capabilities of the DCSS method on several real datasets compared to the most widely used (AE-based) deep clustering methods that have been discussed in Section~\ref{sec:RelatedWork}.

\subsection{Synthetic Data Demonstration}
\begin{figure}[ht]
\centering
\begin{subfigure}{0.24\linewidth}
\centering
\includegraphics[width=\linewidth]{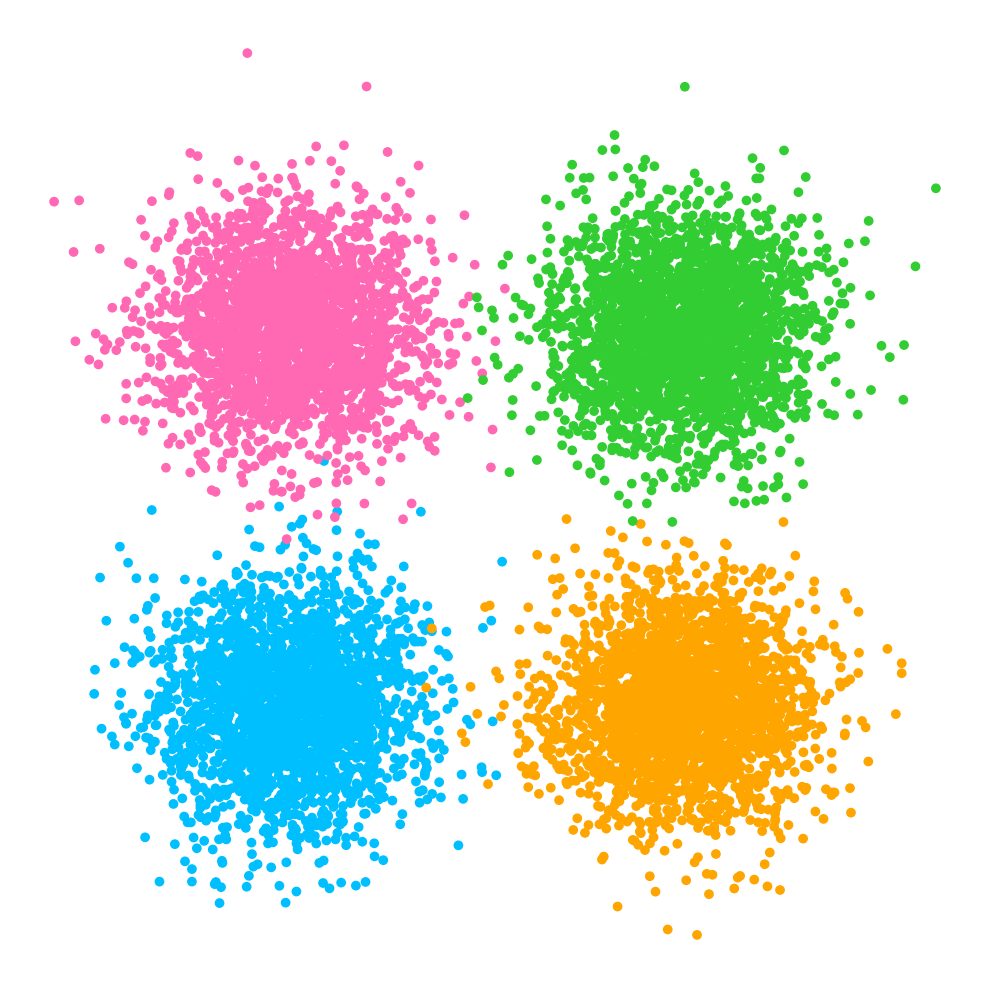}
\caption*{Generated}
\end{subfigure}
\begin{subfigure}{0.24\linewidth}
\centering
\includegraphics[width=\linewidth]{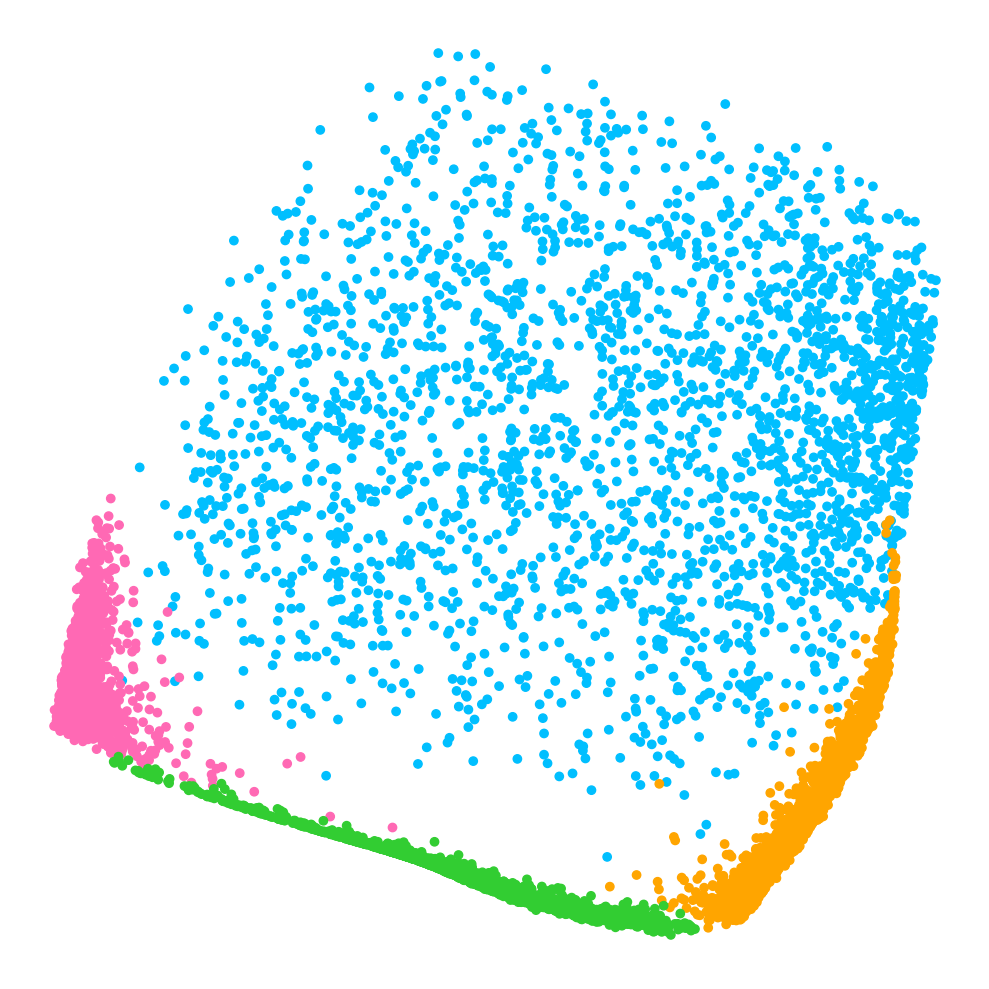}
\caption*{PCA}
\end{subfigure}
\begin{subfigure}{0.24\linewidth}
\centering
\includegraphics[width=\linewidth]{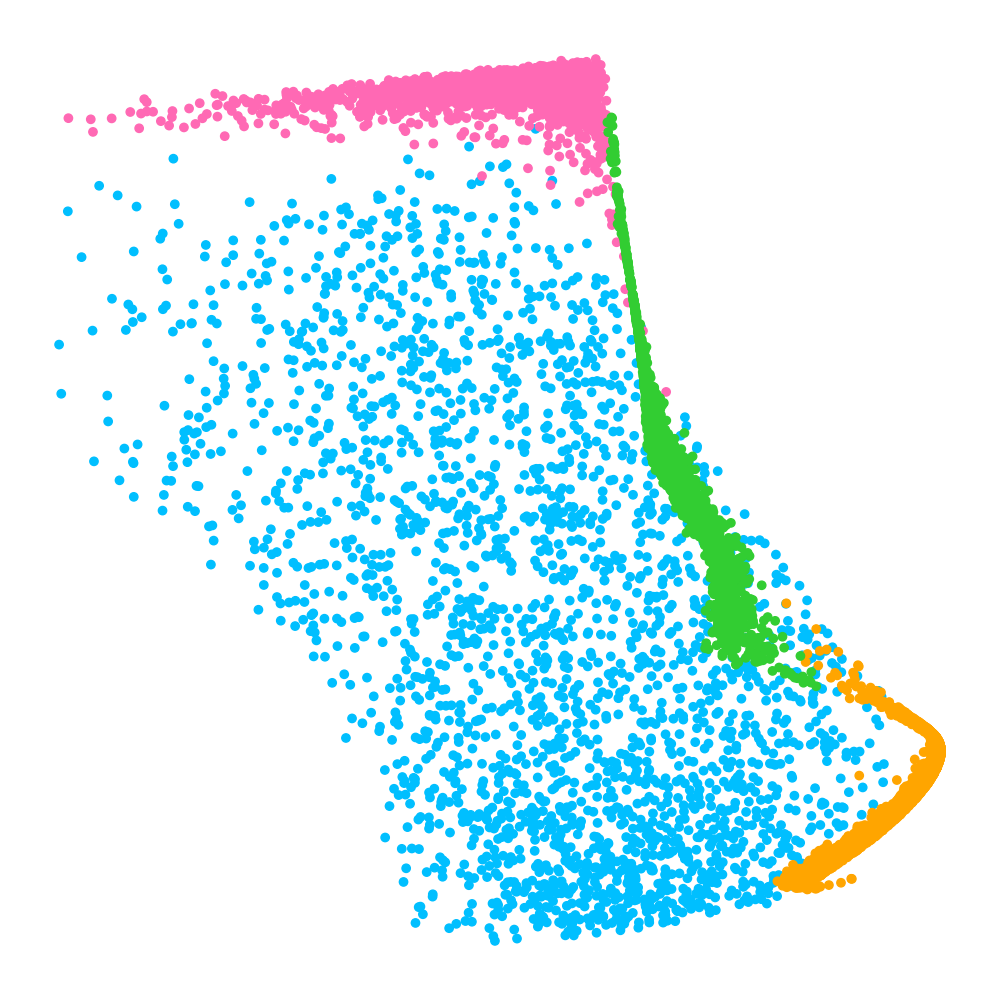}
\caption*{SVD}
\end{subfigure}
\begin{subfigure}{0.24\linewidth}
\centering
\includegraphics[width=\linewidth]{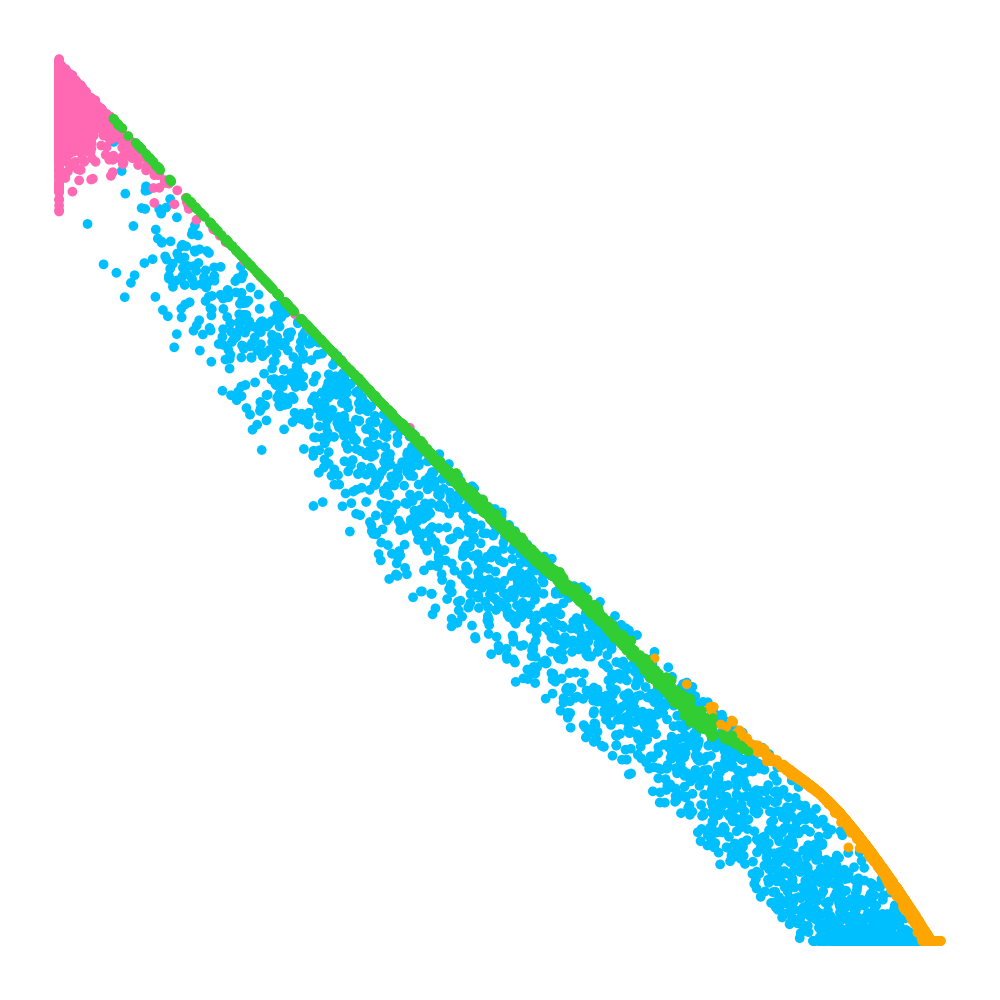}
\caption*{NMF}
\end{subfigure}

\begin{subfigure}{0.24\linewidth}
\centering
\includegraphics[width=\linewidth]{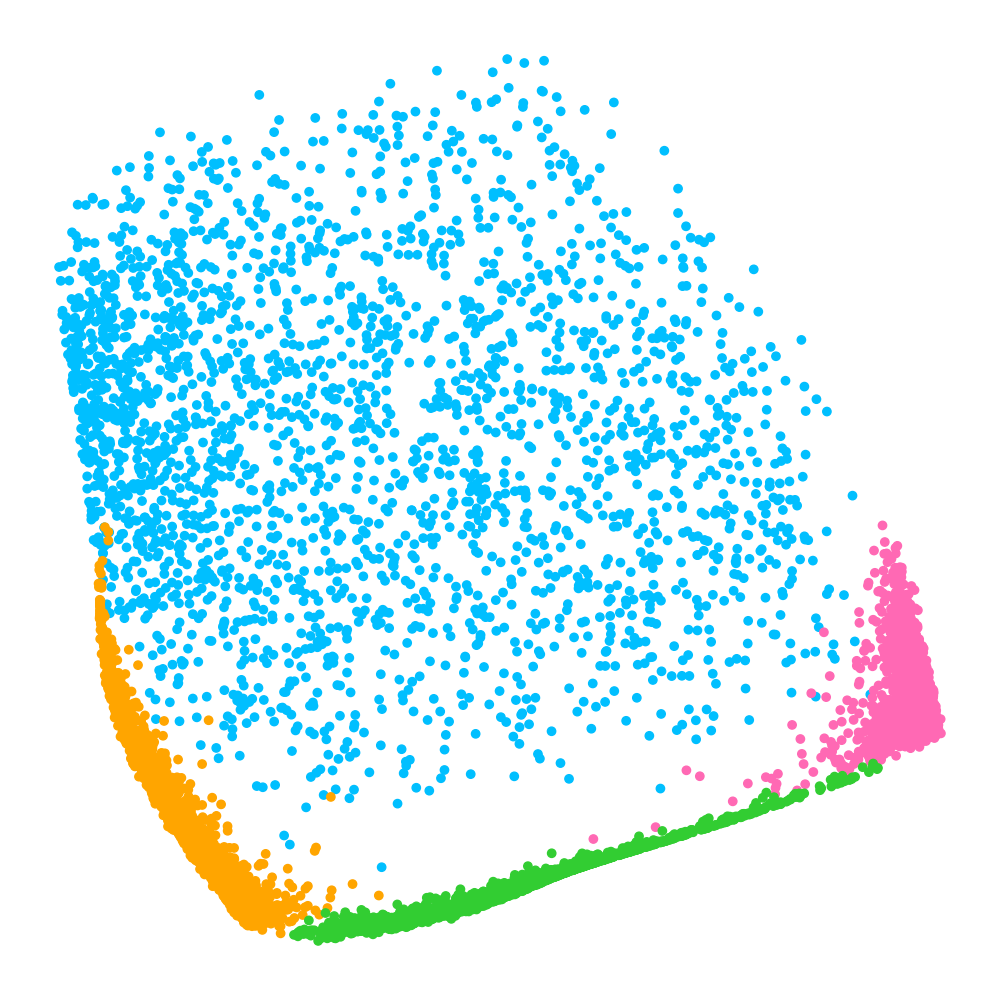}
\caption*{LapEig}
\end{subfigure}
\begin{subfigure}{0.24\linewidth}
\centering
\includegraphics[width=\linewidth]{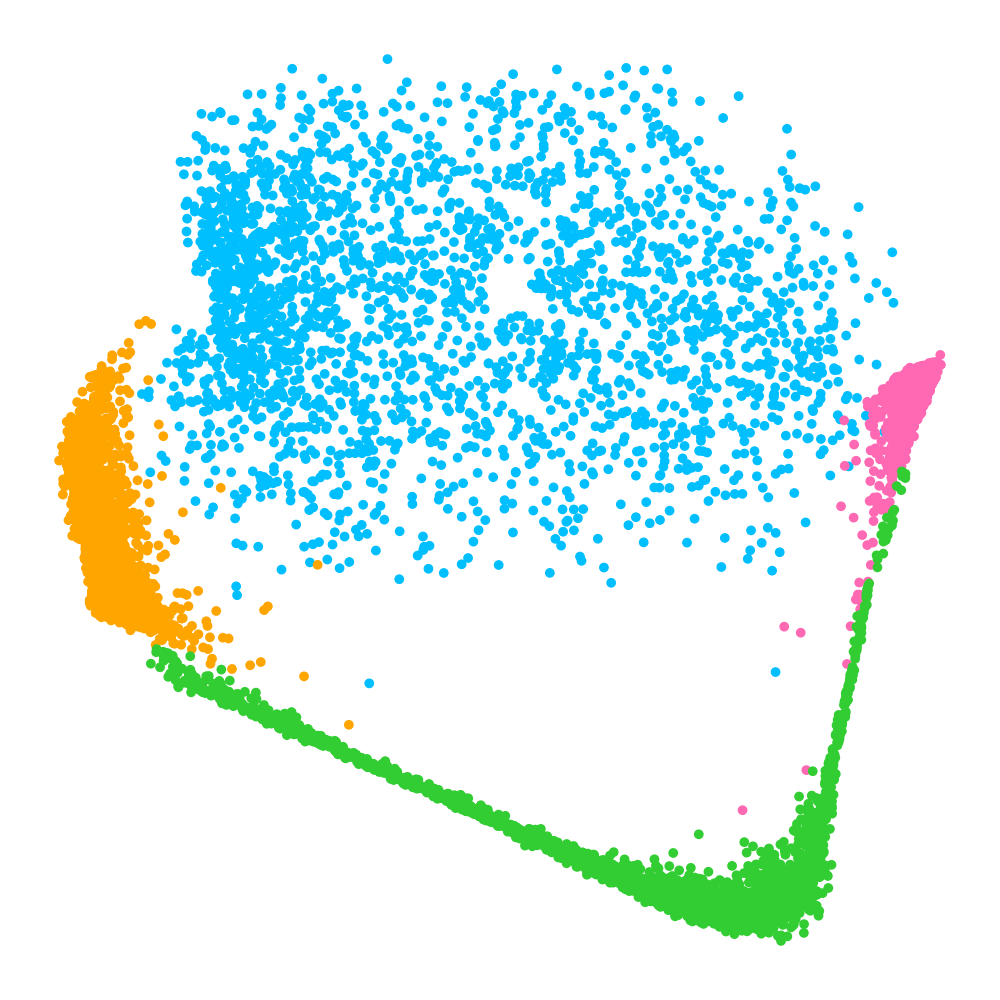}
\caption*{LLE}
\end{subfigure}
\begin{subfigure}{0.24\linewidth}
\centering
\includegraphics[width=\linewidth]{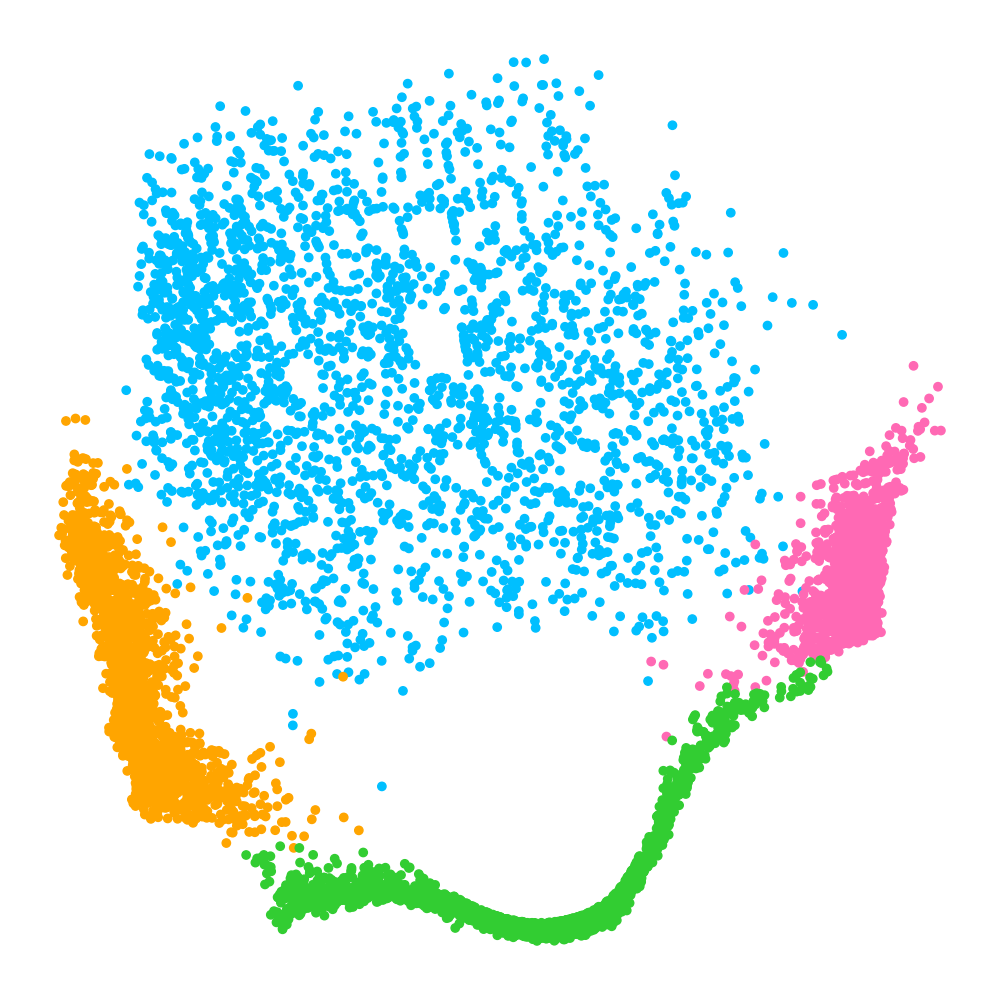}
\caption*{Isomap}
\end{subfigure}
\begin{subfigure}{0.24\linewidth}
\centering
\includegraphics[width=\linewidth]{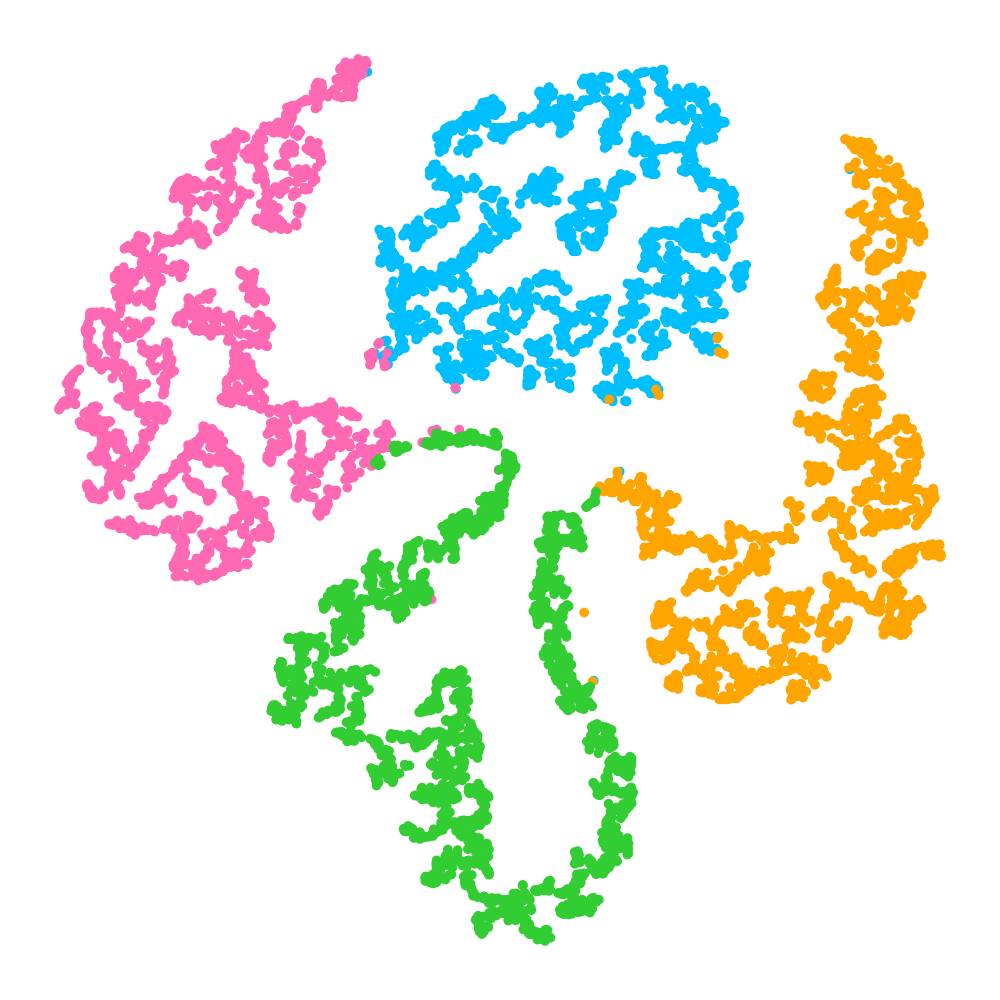}
\caption*{t-SNE}
\end{subfigure}

\begin{subfigure}{0.24\linewidth}
\centering
\includegraphics[width=\linewidth]{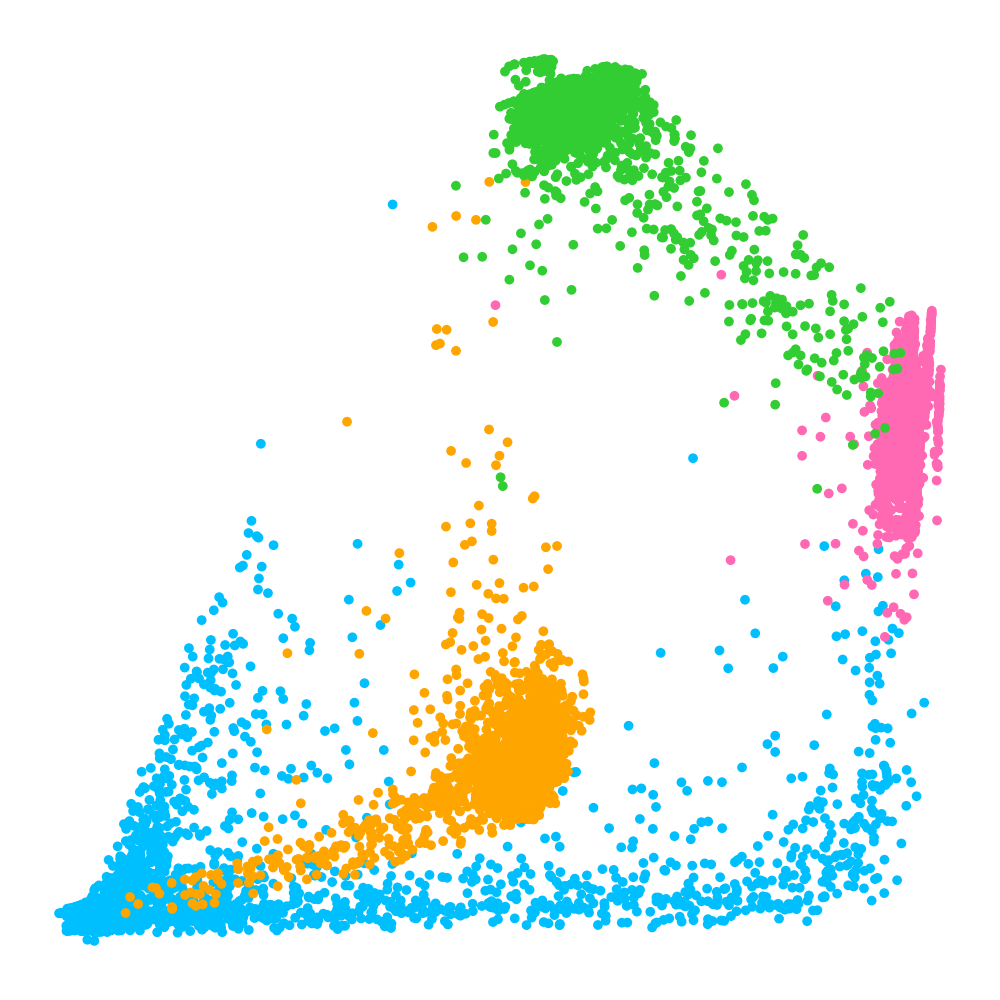}
\caption*{DEC}
\end{subfigure}
\begin{subfigure}{0.24\linewidth}
\centering
\includegraphics[width=\linewidth]{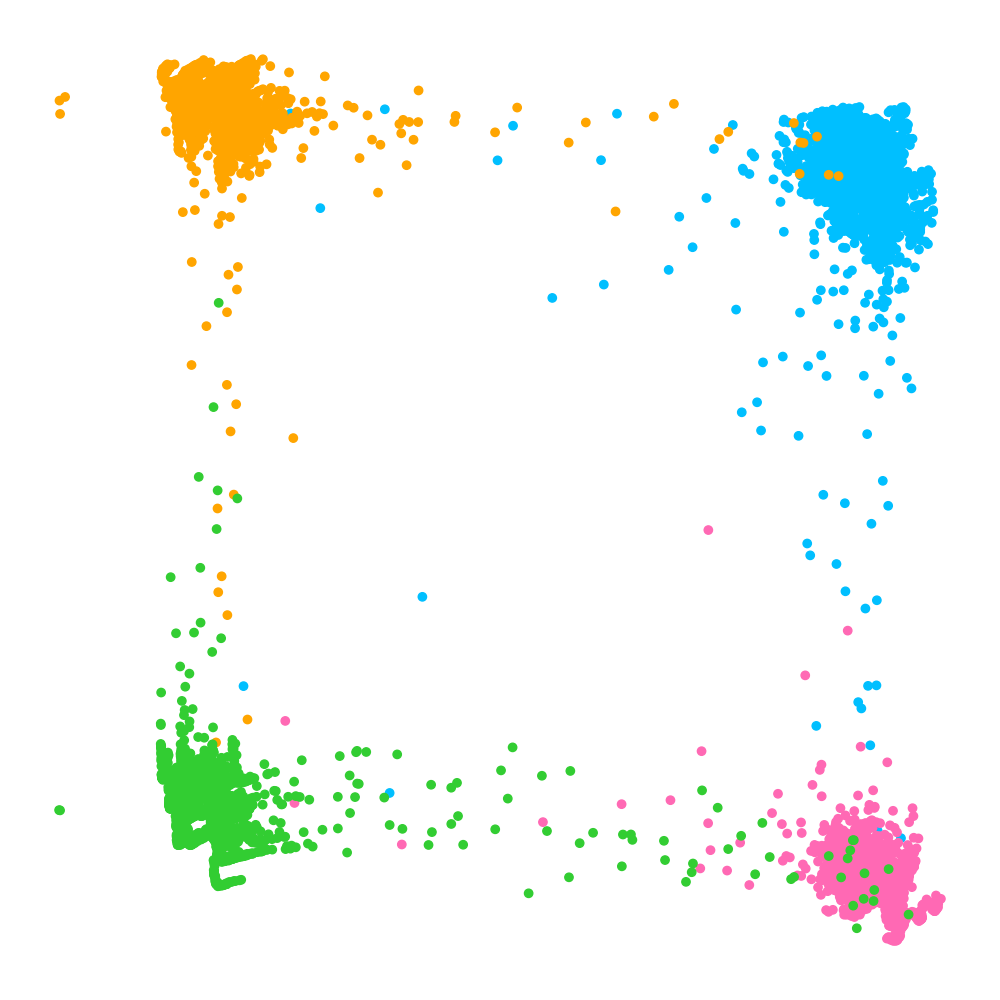}
\caption*{IDEC}
\end{subfigure}
\begin{subfigure}{0.24\linewidth}
\centering
\includegraphics[width=\linewidth]{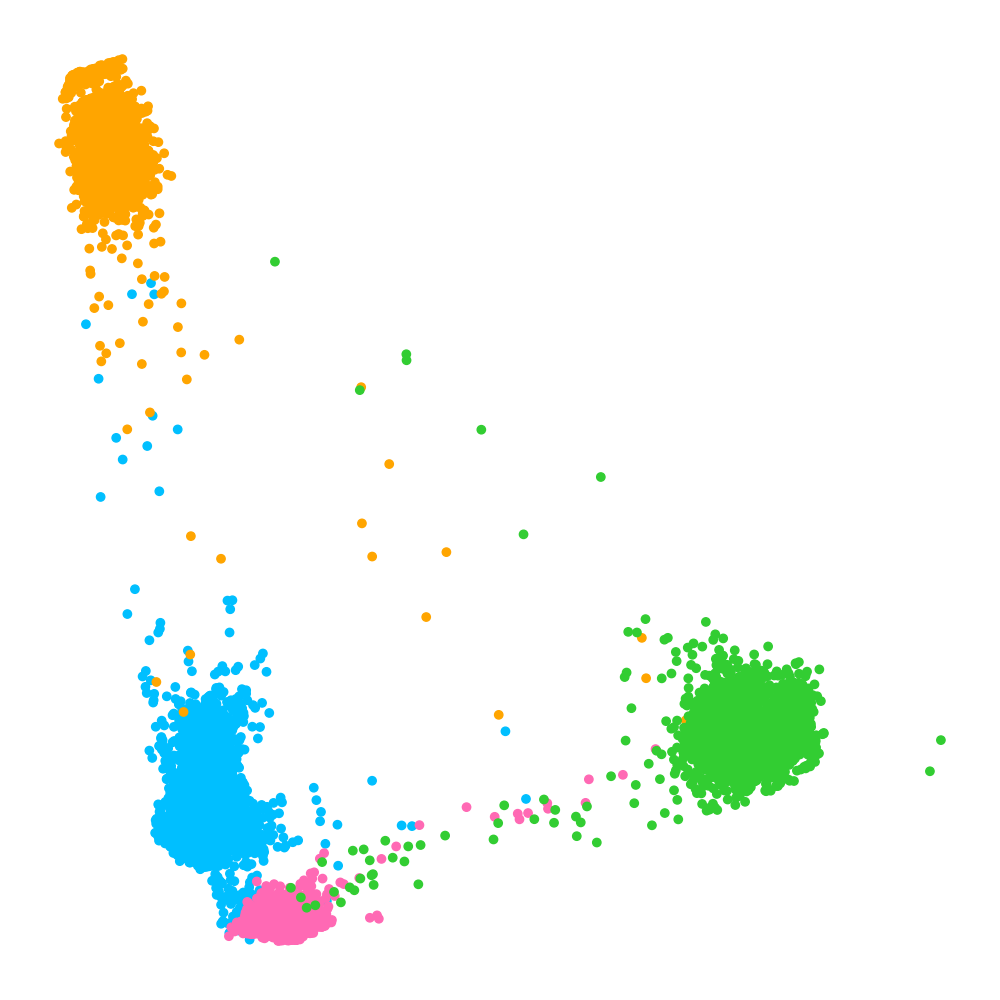}
\caption*{DCN}
\end{subfigure}
\begin{subfigure}{0.24\linewidth}
\centering
\includegraphics[width=\linewidth]{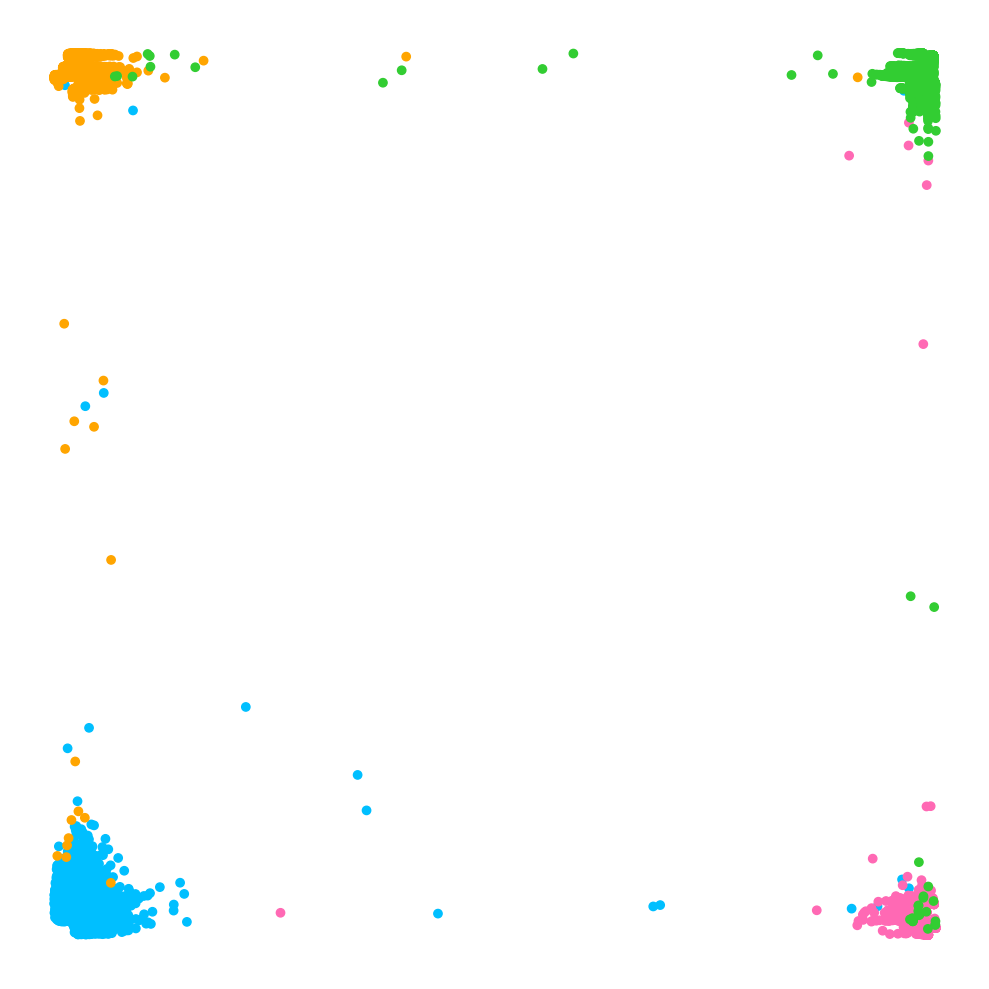}
\caption*{DCSS}
\end{subfigure}
\caption{Synthetic demonstration of the representation learning capabilities of several methods. The generated 2-d dataset (top left) is hidden from the methods. Each method receives as input a 100-d dataset generated by non-linear transformations applied to the original 2-d data and provides a 2-d latent representation of the 100-d dataset, which is presented in the plots. Color indicates the true cluster labels.}
\label{fig:Synthetic-Data}
\end{figure}
We have relied on the synthetic dataset considered in~\cite{yang2017towards} (for testing the DCN method) generated as follows. Let's suppose the observed high dimensional data points $x_i$ exhibit a clear cluster structure in a two-dimensional latent space $Z$, ie. the latent vectors $z_i\in \mathbb{R}^{2}$ form compact and well-separated clusters. Given a latent vector $z_i$, the corresponding observation $x_i$ is generated using the following transformation:
\begin{equation}
x_i = \sigma(U\sigma(Wz_i)),
\end{equation}
where $W \in \mathbb{R}^{10 \times 2}$ and $U \in \mathbb{R}^{100 \times 10}$ are matrices whose entries are sampled from the Normal distribution $\mathcal{N}(0, 1)$, and $\sigma(x)$ is the logistic function that introduces non-linearity into the generation process. Given the observations $x_i$, recovering the original clustering-friendly domain in which $z_i$ resides appears challenging.

We generated a set of latent vectors $z_i$ belonging to four planar clusters, each with 2,500 samples (as shown in the first subfigure of Fig.~\ref{fig:Synthetic-Data}) and we computed the corresponding observations $x_i$. The rest subfigures Fig.~\ref{fig:Synthetic-Data} demonstrate the two-dimensional projections  provided using several dimensionality reduction methods given the observations $x_i$ as input. More specifically, we present the solutions provided by Principal Component Analysis (PCA)~\cite{wold1987principal}, Singular Value Decomposition (SVD), Non-negative Matrix Factorization (NMF)~\cite{lee1999learning}, Local Linear Embeddings (LLE)~\cite{saul2003think}, Isomap~\cite{tenenbaum2000global}, t-SNE~\cite{van2008visualizing} and Laplacian Eigenmap (LapEig), which is the first algorithmic step of the spectral clustering method~\cite{ng2002spectral}. We also considered the deep clustering methods DEC~\cite{xie2016unsupervised}, IDEC~\cite{guo2017improved}, DCN~\cite{yang2017towards}, as well as our proposed DCSS method.

It is evident that the projection methods that do not optimize a clustering loss (first and second rows) have failed to reveal the hidden two-dimensional latent structure. On the contrary, deep clustering methods (bottom row) demonstrated  better performance. Specifically, DEC was able to recover three out of four latent clusters, however the fourth was scattered. DCN was able to reconstruct all of them, but the blue and red clusters are not sufficiently separated. IDEC was able to learn a very informative projection revealing the four cluster structure. Superior are the results of the DCSS method, which was not only able to reveal the four clusters, but also sufficiently maximized their separation.

\subsection{Datasets}
Table~\ref{tab:data} summarizes the benchmark datasets that we used for experimental evaluation, which vary in size $n$, dimensions $d$, number of clusters $k$, complexity, data type, and domain of origin. The subsequent paragraphs offer a more comprehensive overview of the datasets we employed and outline the preprocessing procedures we implemented for each of these datasets.

The datasets used in this study include the Pendigits (PEN), EMNIST MNIST (E-MNIST), and EMNIST Balanced Digits (BD). These datasets consist of handwritten digits categorized into ten classes, each representing digits from $0$ to $9$. It is worth noting that the EMNIST dataset constitutes an extended and more challenging version of the MNIST dataset~\cite{lecun-mnisthandwrittendigit-2010}. Both the E-MNIST and BD datasets comprise images with a resolution of $28\times28$ pixels. In contrast, Pendigits data instances are represented by $16$-dimensional vectors containing pixel coordinates.

In addition, the EMNIST Balanced Letters (BL) dataset is included, featuring handwritten letters in both uppercase and lowercase forms, with a resolution of $28\times28$ pixels. The BL dataset has been divided into three mutually exclusive subsets. The first subset contains the letters A to J, the second includes the letters K to T, and the last subset contains the remaining letters U to Z. The first two subsets comprise $28000$ data points each, distributed across $10$ clusters, while the last subset consists of $16800$ data points and $6$ clusters. The Human Activity Recognition with Smartphones (HAR) dataset was also considered. This dataset consists of data collected from the accelerometer and gyroscope sensors of smartphones, sampled during a human activity. Specifically, each record in the dataset is a $560$ feature vector with time and frequency domain variables. In addition, HAR consists of $6$ classes of human activities which are the following: walking, walking upstairs, walking downstairs, sitting, standing, laying. Furthermore, the Waveform-v1 (WVF-v1) dataset was included which consists of 3 classes of generated waves with $5000$ datapoints. Each class is generated from a combination of 2 of 3 `base' waves. Finally, we also report results for the synthetic dataset described in the previous subsection.

In all datasets, we used min-max normalization as a prepossessing step, to map the attributes of each data point to the $[0, 1]$ interval to prevent attributes with large ranges from dominating the distance calculations and avoid numerical instabilities in the computations~\cite{celebi2013comparative}. 

\begin{table}[t]\footnotesize
\centering
\caption{The datasets used in our experiments. $N$ is the number of data instances, $d$ is the dimensionality, and $k$ denotes the number of clusters.}
\label{tab:data}
\begin{tabular}{llrrrc}
\toprule
\textbf{Dataset}&\textbf{Type}&{$N$}&\textbf{$d$}&\textbf{$k$}&\textbf{Source}\\
\midrule
EMNIST Balanced Digits&Image&28000&$28\times28$&10&\cite{cohen2017emnist}\\
EMNIST MNIST&Image&70000&$28\times28$&10&\cite{cohen2017emnist}\\
EMNIST Balanced Letters (A-J)&Image&28000&$28\times28$&10&\cite{cohen2017emnist}\\
EMNIST Balanced Letters (K-T)&Image&28000&$28\times28$&10&\cite{cohen2017emnist}\\
EMNIST Balanced Letters (U-Z)&Image&16800&$28\times28$&6&\cite{cohen2017emnist}\\
HAR&Tabular&10299&560&6&\cite{har}\\
Pendigits&Tabular&10992&16&10&\cite{Dua:2019}\\
Waveform-v1&Tabular&5000&21&3&\cite{Dua:2019}\\
Synthetic&Tabular&10000&2&4&\cite{yang2017towards}\\
\bottomrule
\end{tabular}
\end{table}

\subsection{Neural Network Architectures}
Determining optimal architectures and hyperparameters through cross-validation is not feasible in unsupervised learning problems. Therefore, we opt for commonly used architectures for the employed neural network models while avoiding dataset-specific tuning. Regarding tabular data, our approach involves adopting a well-established architecture, which consists of fully connected layers~\cite{van2009learning,xie2016unsupervised}. The specific AE architecture that we used is the following:
\begin{center}
$x_{d}\rightarrow$
Fc$_{500}\rightarrow$
Fc$_{500}\rightarrow$
Fc$_{2000}\rightarrow$
Fc$_{m}\rightarrow$
Fc$_{2000}\rightarrow$
Fc$_{500}\rightarrow$
Fc$_{500}\rightarrow$
$\hat{x}_{d}$, 
\end{center}
where Fc$_{m}$ stands for fully connected layer with $m$ neurons and $x_{d}$ represents a d-dimensional data vector.

In what concerns image datasets, convolutional Neural Networks (CNNs) have demonstrated superior effectiveness in capturing semantic visual features. Consequently, we exploit a convolutional-deconvolutional AE to learn the embeddings for the image datasets. The AE architecture consists of three convolutional layers (encoder), one fully connected layer (emdedding layer), and three deconvolutional layers (decoder)~\cite{guo2018deep,ren2020deep,guo2021deep}. More specifically, the architecture is the following:
\begin{center}
$x_{28 \times 28}\rightarrow$
Conv$^{5}_{32}\rightarrow$
Conv$^{5}_{64}\rightarrow$
Conv$^{3}_{128}\rightarrow$
Fc$_{m}\rightarrow$
Deconv$^{3}_{128}\rightarrow$
Deconv$^{5}_{64}\rightarrow$
Deconv$^{5}_{32}\rightarrow$
$\hat{x}_{28 \times 28}$,
\end{center}
where Conv$^{a}_{b}$ (Deconv$^{a}_{b}$) denote a convolutional (deconvolutional) layer with an $a \times a$ kernel and $b$ filters, while the stride always set to 2.

In the above encoder-decoder networks, the ReLU activation function is used~\cite{nair2010rectified}, except for the embedded layer of the AE, where the Hyperbolic Tangent (tanh) function is used. Weights and biases are initialized using the He initialisation method~\cite{he2015delving}. 

As already mentioned, in what concerns the clustering network, a Radial Basis Function (RBF) model has been selected with the number of hidden units equal to the number of clusters. The outputs of the RBF units are fed to a $K$-output softmax activation function that provides the cluster assignment probabilities. After the AE pre-training, we initialize the centers of the RBF layer by using the $k$-means algorithm in the embedded space, while we initialize $\sigma$ to a small positive value. The temperature parameter $T$ of the softmax was set equal to $T=20$.

\begin{table}[ht]\footnotesize

\centering
\caption{Performance results of the compared clustering methods.}
\label{tab:results}

\label{results}
\begin{tabular}{ll|rrrrrr}
\toprule
\hline
 & & \multicolumn{6}{c}{\textbf{Method}} \\ \hline
\textbf{Dataset} & \textbf{Measure} & $k$-means & AE $\texttt{+}$ $k$-means & DCN & DEC & IDEC & DCSS  \\ \hline

 \multirow{2}{*}{BD} 
 & NMI & 0.48 & $0.72 \pm $0.01 & $0.75 \pm $0.02 & $0.80 \pm $0.03 & $0.82 \pm $0.01 & $\mathbf{0.86} \pm $0.04 \\
 & ARI & 0.36 & $0.65 \pm $0.02 & $0.63 \pm $0.05 & $0.75 \pm $0.06 & $0.77 \pm $0.01 & $\mathbf{0.80} \pm $0.07 \\ \hline

\multirow{2}{*}{BL (A-J)} 
& NMI & 0.35 & $0.65 \pm $0.02 & $0.68 \pm $0.03 & $0.77 \pm $0.03 & $0.77 \pm $0.03 & $\mathbf{0.80} \pm $0.02 \\
& ARI & 0.25 & $0.55 \pm $0.02 & $0.56 \pm $0.05 & $0.69 \pm $0.06 & $0.69 \pm $0.05 & $\mathbf{0.72} \pm $0.04 \\ \hline

\multirow{2}{*}{BL (K-T)} 
& NMI & 0.51 & $0.73 \pm $0.02 & $0.77 \pm $0.03 & $0.84 \pm $0.01 & $0.84 \pm $0.02 & $\mathbf{0.90} \pm $0.02 \\
& ARI & 0.43 & $0.67 \pm $0.04 & $0.69 \pm $0.06 & $0.81 \pm $0.02 & $0.81 \pm $0.04 & $\mathbf{0.87} \pm $0.03 \\ \hline

\multirow{2}{*}{BL (U-Z)} 
& NMI & 0.47 & $0.64 \pm $0.01 & $0.64 \pm $0.02 & $0.68 \pm $0.02 & $0.67 \pm $0.02 & $\mathbf{0.71} \pm $0.04 \\
& ARI & 0.41 & $0.60 \pm $0.02 & $0.56 \pm $0.04 & $0.65 \pm $0.03 & $0.63 \pm $0.02 & $\mathbf{0.68} \pm $0.06 \\ \hline
 
\multirow{2}{*}{E-MNIST} 
& NMI & 0.48 & $0.75 \pm $0.01 & $0.83 \pm $0.03 & $0.84 \pm $0.03 & $0.85 \pm $0.02 & $\mathbf{0.88} \pm $0.04 \\
& ARI & 0.36 & $0.69 \pm $0.01 & $0.78 \pm $0.04 & $0.80 \pm $0.04 & $0.81 \pm $0.03 & $\mathbf{0.83} \pm $0.06 \\ \hline

\multirow{2}{*}{HAR} 
& NMI & 0.59 & $0.67 \pm $0.06 & $0.77 \pm $0.01 & $0.74 \pm $0.06 & $0.74 \pm $0.10 & $\mathbf{0.81} \pm $0.06 \\
& ARI & 0.46 & $0.60 \pm $0.09 & $0.70 \pm $0.02 & $0.66 \pm $0.08 & $0.65 \pm $0.12 & $\mathbf{0.74} \pm $0.09 \\ \hline

\multirow{2}{*}{PEN} 
& NMI & 0.69 & $0.68 \pm $0.02 & $0.73 \pm $0.03 & $0.73 \pm $0.03 & $0.75 \pm $0.03 & $\mathbf{0.78} \pm $0.02 \\
& ARI & 0.56 & $0.59 \pm $0.04 & $0.62 \pm $0.05 & $0.62 \pm $0.05 & $0.65 \pm $0.04 & $\mathbf{0.68} \pm $0.04 \\ \hline

\multirow{2}{*}{WVF-v1} 
& NMI & 0.74 & $0.84 \pm $0.13 & $0.95 \pm $0.08 & $0.93 \pm $0.12 & $0.89 \pm $0.09 &  $\mathbf{1.00} \pm $0.00\\
& ARI & 0.70 & $0.87 \pm $0.13 & $0.95 \pm $0.09 & $0.93 \pm $0.14 & $0.91 \pm $0.08 &  $\mathbf{1.00} \pm $0.00 \\ \hline

\multirow{2}{*}{Synthetic} 
& NMI & 0.82 & $0.72 \pm $0.14 & $0.79 \pm $0.09 & $0.91 \pm $0.08 & $0.90 \pm $0.10 & $\mathbf{0.93} \pm $0.10 \\
& ARI & 0.83 & $0.70 \pm $0.19 & $0.75 \pm $0.14 & $0.92 \pm $0.12 & $0.91 \pm $0.11 & $\mathbf{0.93} \pm $0.10 \\ \hline


\bottomrule
\end{tabular}
\end{table}

\subsection{Evaluation}
It is important to mention that since clustering is an unsupervised problem, we ensured that all algorithms were unaware of the true clustering of the data. In order to evaluate the results of the clustering methods, we use standard external evaluation measures, which assume that ground truth clustering is available~\cite{rendon2011internal}. For all algorithms, the number of clusters is set to the number of ground-truth categories and assumes ground truth that cluster labels coincide with class labels. 
The first evaluation measure is the \emph{Normalized Mutual Information} (NMI)~\cite{estevez2009normalized} defined as:
\begin{equation}
\label{eq:NMI}
NMI(Y, C) = \frac{2 \times I(Y, C)}{H(Y) + H(C)},
\end{equation}
where $Y$ denotes the ground-truth labels, $C$ denotes the clusters labels, $I(\cdot)$ is the mutual information measure and $H(\cdot)$ the entropy. The second metric used is the \emph{Adjusted Rand Index} (ARI)~\cite{hubert1985comparing,chacon2022minimum}, which is a corrected for chance version of the Rand Index~\cite{rand1971objective} that measures the degree of overlap between two partitions defined as:
\begin{equation}
\label{eq:ARI}
ARI(Y, C) = \frac{RI(Y, C) - \mathbb{E}[RI(Y, C)]}{max\{RI(Y, C)\} - \mathbb{E}[RI(Y, C)]},
\end{equation}
where $RI(\cdot)$ denotes the Rand Index and $\mathbb{E}[\cdot]$ is the expected value.

\subsection{Experimental Setup and Results}

We have conducted a comprehensive performance analysis of the proposed DCSS method in comparison to well-studied deep clustering methods such as DCN~\cite{yang2017towards}, DEC~\cite{xie2016unsupervised}, and IDEC~\cite{guo2017improved}. These methods are designed to facilitate the learning of a cluster-friendly embedded space as also happens with our approach.
Furthermore, we have evaluated the performance of $k$-means~\cite{lloyd1982least} both in the original space and in the embedded space (AE+$k$-means). The comparison with the latter approach quantifies the performance improvements achieved through the utilization of AE in the clustering procedure. At this point, it should be noted that for a fair comparison between the deep clustering methods, we used the same model architectures for all the methods, since we observed improved clustering results compared to those proposed in the original papers.

In experiments involving $k$-means, we initialized the algorithm $100$ times and retained the clustering solution with the lowest mean sum of squares error. For the remaining methods, which integrate an AE model in the clustering procedure, we conducted each experiment $10$ times. In the context of deep clustering methods, an AE pre-training phase (ignoring the clustering loss) took place. For image datasets, we pretrained the AE for 100 epochs with a learning rate of $1 \times 10^{-3}$, while for tabular datasets, we extended the pre-training to 1000 epochs with a learning rate of $5 \times 10^{-4}$. During the pre-training phase, a small $L_2$ regularization of $1 \times 10^{-5}$ was applied. In the training phase, the deep clustering models were trained for 100 epochs with a learning rate of $5 \times 10^{-4}$ and no regularization penalty. A fixed batch size of $256$  was considered and the Adam optimizer~\cite{kingma2014adam} with the default settings of $\beta_1 = 0.9$ and $\beta_2 = 0.999$ was used in both the pretraining and training phases. Additionally, there are several methodologies to tune the non-clustering (eq.~\ref{eq:rec_loss}) with the clustering loss (eq.~\ref{eq:cluster_loss})~\cite{aljalbout2018clustering} during training. We choose the most simplistic and typical approach by setting our hyper-parameters to a small values that balances the two losses. Specifically, we used $\lambda_1 = 0.01$ and $\lambda_2 = 0.01$. Finally, to initialize the centers of the RBF layer, we apply $k$-means in the embeddings of the pretrained AE, while $\sigma$ was initialized to to a small positive value.

In Table~\ref{tab:results}, we present the average performance in terms of NMI and ARI along with the standard deviation for each method and dataset. As anticipated, the clustering performance of $k$-means improves when projecting the data to low-dimensional embedded space, as shown in the AE+$k$-means compared to $k$-means results. In general, the performance of DEC is better than that of DCN in all datasets except HAR and WVF-v1. At the same time, we can observe that IDEC slightly outperforms the DEC method in most cases, indicating that some improvement might be obtained by using the decoder part in the clustering optimization procedure. Finally, the results demonstrate the superiority of the proposed DCSS method across all datasets. Concerning the NMI measure, there is a significant improvement ranging from 0.03 to 0.07. In addition, the ARI measure shows an improvement ranging from 0.03 to 0.06. These results indicate that soft silhouette serves as a more suitable deep clustering objective function capable of providing cluster-friendly representations. This is attributed to its optimization approach, which aims for compact and well-separated clusters.

\begin{figure}[ht]
\centering
\resizebox{0.7\linewidth}{!}{
\begin{subfigure}{0.48\linewidth}
\centering
\includegraphics[width=\linewidth]{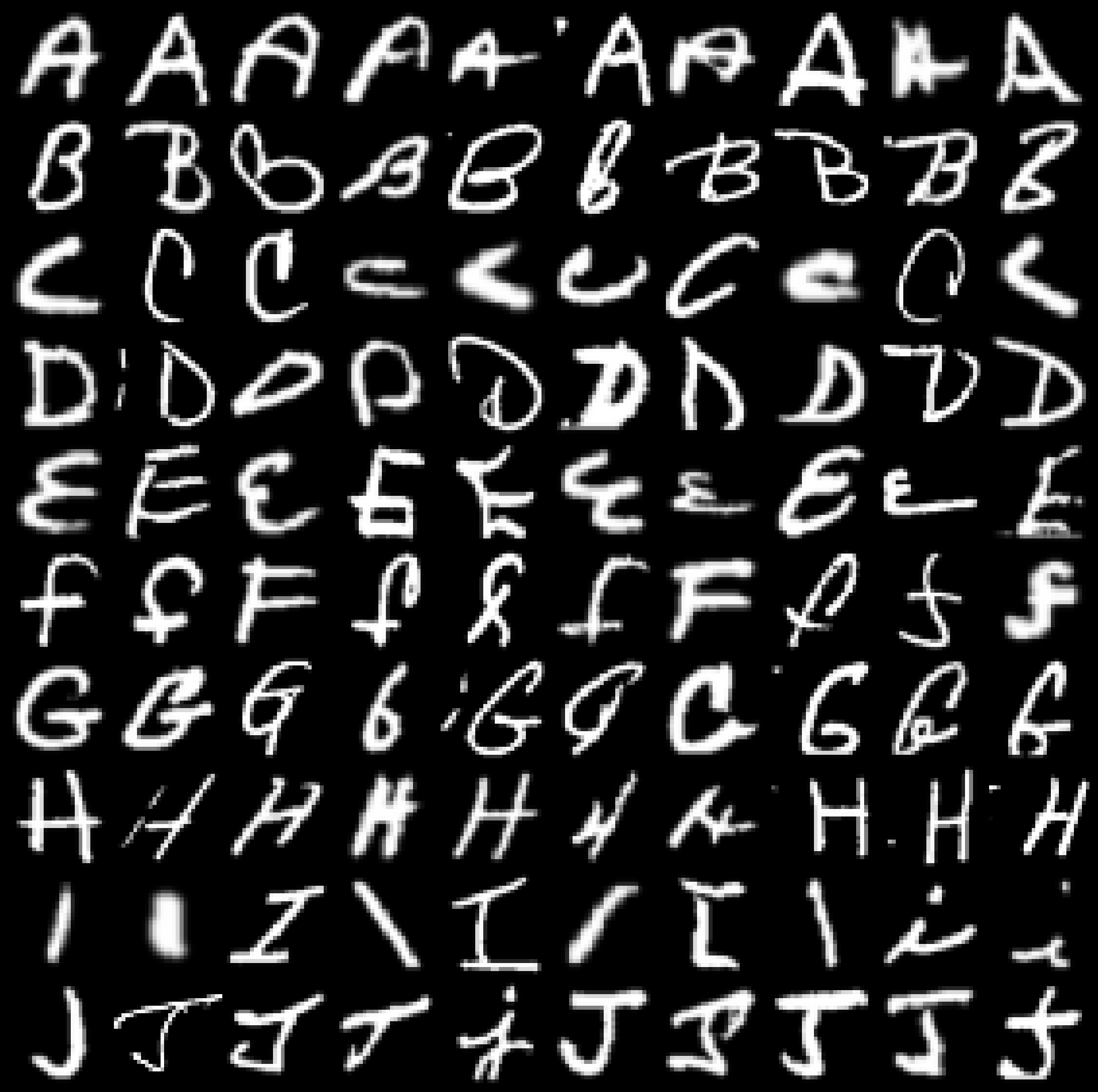}
\caption*{BL (A-J)}
\end{subfigure}
\begin{subfigure}{0.48\linewidth}
\centering
\includegraphics[width=\linewidth]{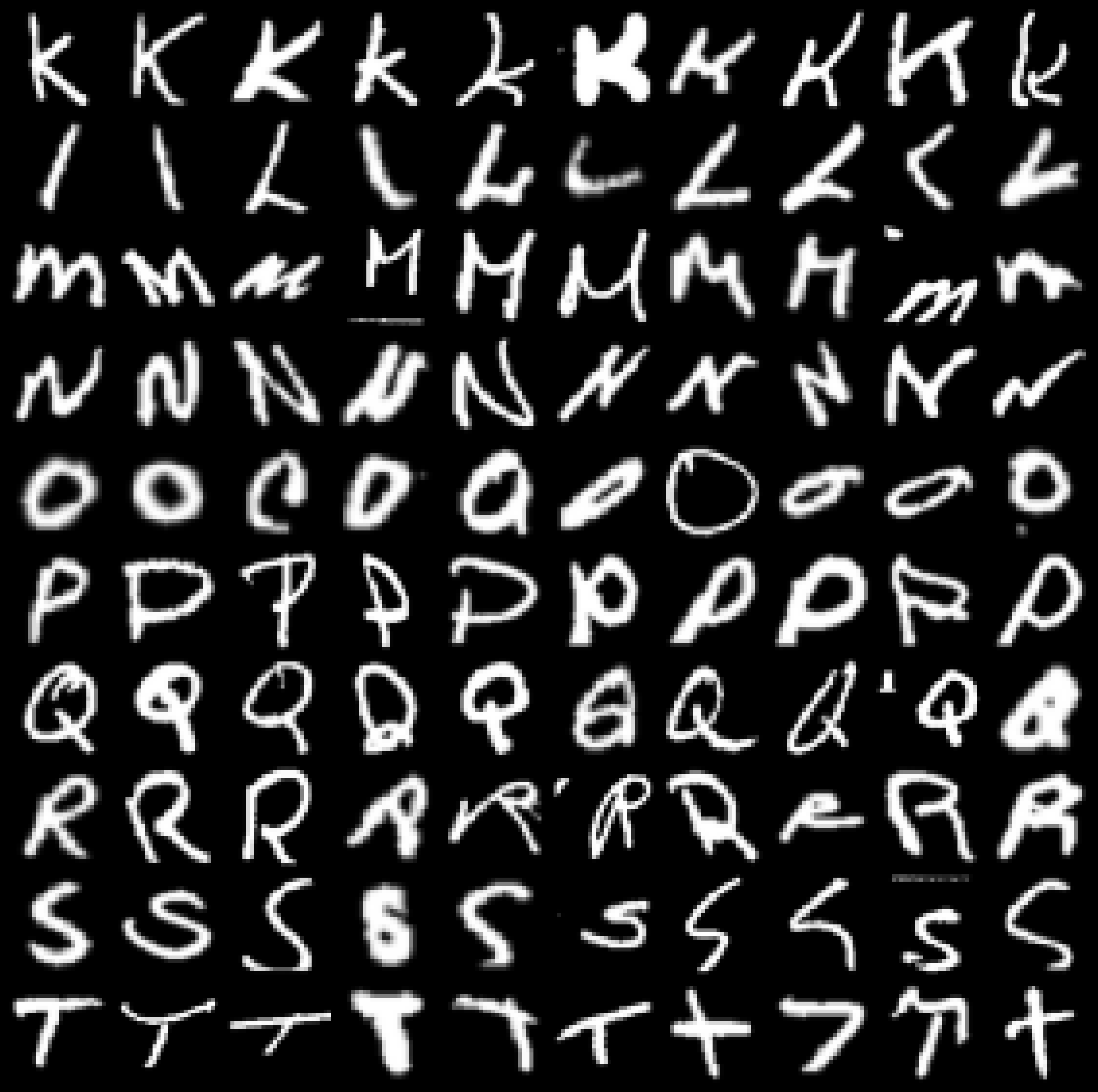}
\caption*{BL (K-T)}
\end{subfigure}
}

\resizebox{0.7\linewidth}{!}{
\begin{subfigure}{0.48\linewidth}
\centering
\includegraphics[width=\linewidth]{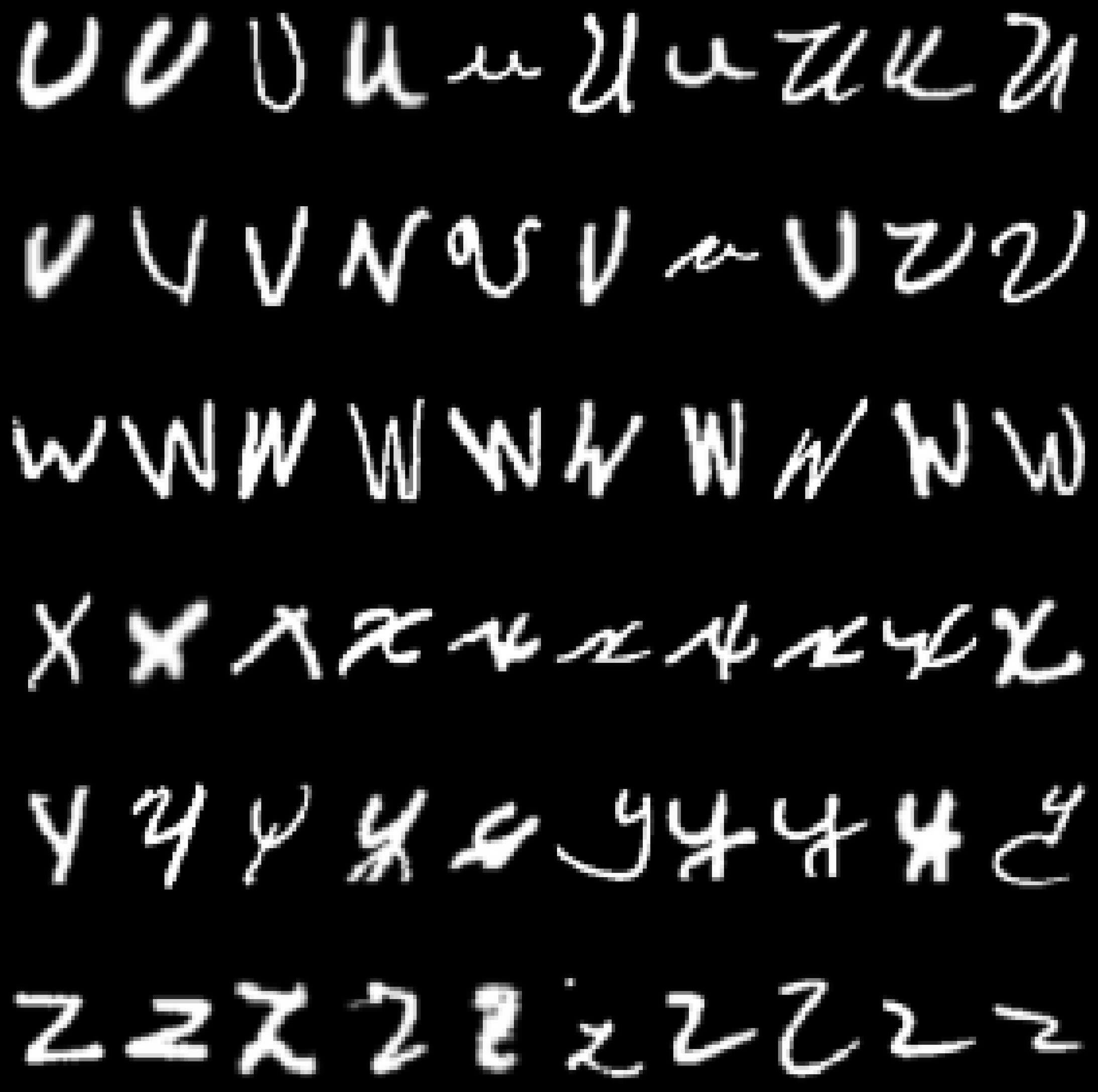}
\caption*{BL (U-Z)}
\end{subfigure}
\begin{subfigure}{0.48\linewidth}
\centering
\includegraphics[width=\linewidth]{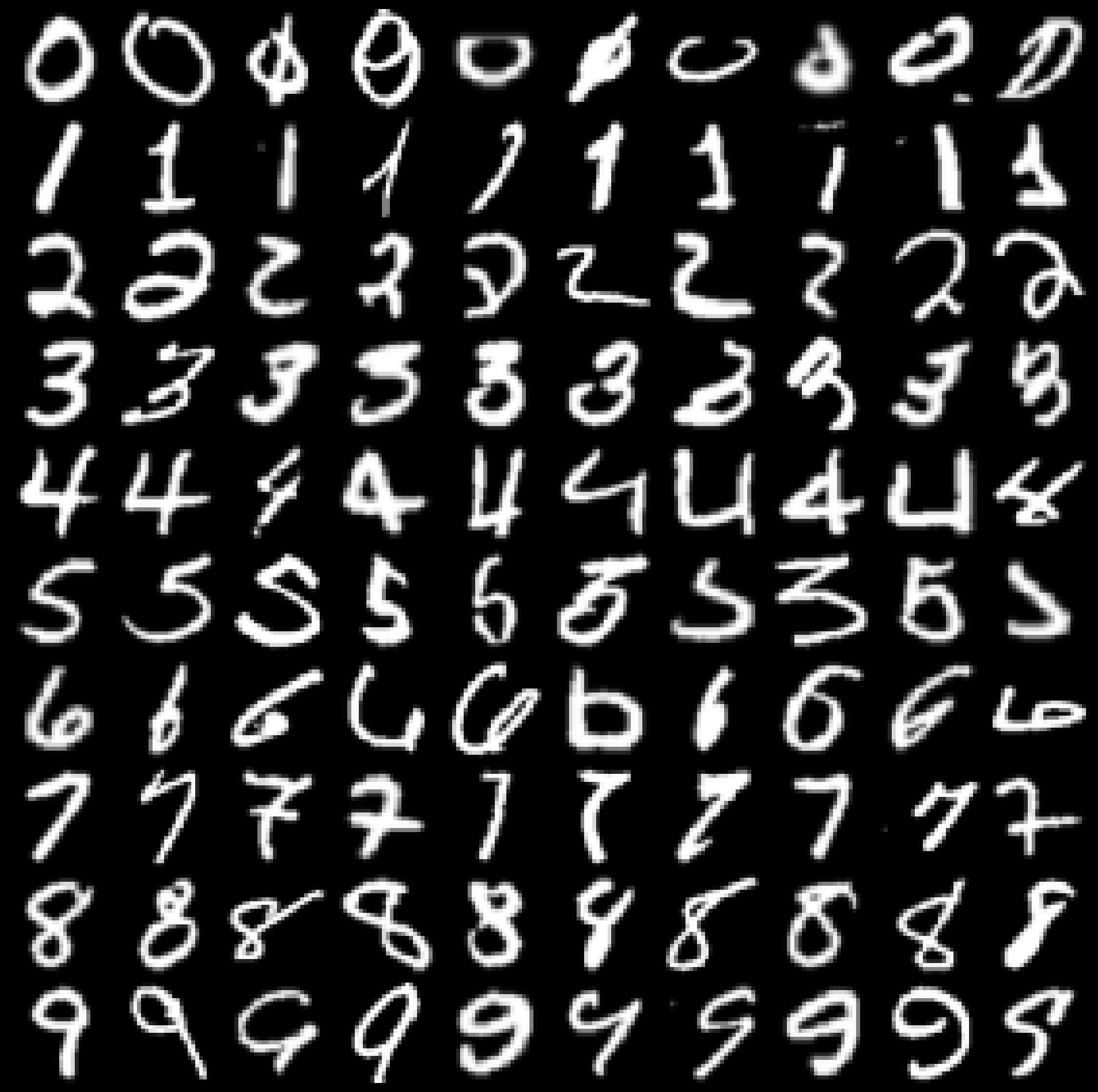}
\caption*{E-MNIST}
\end{subfigure}
}

\caption{Image clustering results on various datasets. In each sub-figure, rows correspond to different clusters. In each row the images are presented from left to right with decreasing cluster membership probability.}
\label{fig:Image-Clustering}
\end{figure}

In Figure~\ref{fig:Image-Clustering}, we present image clustering results using the DCSS method on datasets BL (A-J), BL (K-T), BL (U-Z), and E-MNIST. Images in the same row are assigned to the same cluster and are placed with decreasing cluster assignment probability, progressing from the leftmost (high probability) to the rightmost (low probability) columns. It is obvious that more representative images are assigned higher probability values. 

\section{Conclusions}
\label{sec:Conclusion}
In this paper, we have proposed soft silhouette, an extension of the widely used silhouette score that accounts for probabilistic clustering assignments. Next, we have considered soft silhouette as a differentiable clustering objective function and propose the DCSS deep clustering methodology that constitutes an autoencoder-based approach suitable for optimizing the soft silhouette score. The DCSS method guides the learned latent representations to form both compact and well-separated clusters. This property is crucial in real-world applications, as targeting both compactness and separability ensures that the resulting clusters are not only densely packed but also distinct from each other. 

The proposed method has been tested and compared with well-known deep clustering methods on various benchmark datasets, yielding very satisfactory results. The experimental study indicates that soft silhouette constitutes as a more suitable deep clustering objective function capable of enhancing the learned representations of the embedded space for clustering purposes.

There are several directions for future work, such as improving the clustering results using data augmentation techniques since it is adopted as an effective strategy for enhancing the learned representations~\cite{guo2018deep,deng2023strongly}. In addition, more sophisticated models and training methodologies can be used, such as ensemble models~\cite{affeldt2020spectral} or adversarial learning~\cite{yang2020adversarial}. It is also possible to modify the learning procedure to incorporate self-paced learning~\cite{kumar2010self}, since learning the 'easier' data first is expected to improve the clustering results~\cite{li2018discriminatively,guo2019adaptive,zhang2023self}. However, our major focus will be on extending the DCCS algorithm for estimating the number of clusters by exploiting unimodality tests as happens in the dip-means~\cite{kalogeratos2012dip} and DIPDECK~\cite{leiber2021dip} algorithms. 



\section*{Code Availability}
An (early) version of the code and relevant experiments will soon be available in the following github repository: \href{https://github.com/gvardakas/Soft-Silhouette}{https://github.com/gvardakas/Soft-Silhouette}. 

	
\bibliographystyle{unsrt}
\bibliography{Bibliography.bib}

\end{document}